\documentclass{article}

\PassOptionsToPackage{numbers, compress}{natbib}

\usepackage[main, final]{neurips_2026}

\usepackage[utf8]{inputenc} 
\usepackage[T1]{fontenc}    
\usepackage{hyperref}       
\usepackage{url}            
\usepackage{booktabs}       
\usepackage{amsfonts}       
\usepackage{nicefrac}       
\usepackage{microtype}      
\usepackage{xcolor}         
\usepackage{amsmath, amssymb, amsthm}
\usepackage{enumitem}
\usepackage{graphicx}       
\usepackage{array}          
\usepackage{caption}        
\usepackage{wrapfig}        
\graphicspath{{fig/}}       

\newtheorem{proposition}{Proposition}

\newtheorem{lemma}{Lemma}

\title{Rethinking Contrastive Targets in Radiology Language-Image Pretraining}

\author{%
  Yunhe Gao,\quad Ashwin Kumar,\quad Jiaming Liu,\quad Chong Wang, \\
  \textbf{Maya Varma,\quad Fuying Wang,\quad Stefania L.~Moroianu,\quad Louisa Fay,} \\
  \textbf{Sergios Gatidis,\quad Akshay S.~Chaudhari,\quad Curtis Langlotz} \\[0.5em]
  Center for Artificial Intelligence in Medicine and Imaging (AIMI) \\
  Stanford University
}

\begin{document}

\maketitle

\begin{abstract}
CLIP-style contrastive image--text pretraining is widely used in radiology data, where each image is paired with one report and every other report in the batch is treated as a negative. Under this recipe, two failures appear, neither of which is systematically addressed by existing contrastive learning theory: zero-shot evaluation peaks then degrades while training loss keeps decreasing, and training fails to learn from large batches. We hypothesize the cause is structural rather than radiology-specific: many unpaired reports share clinical findings with a given image, so CLIP's paired-identity target is systematically biased, not merely noisy. To test this, we construct \emph{CLEVR-Radiology}, a synthetic testbed that preserves only this overlapping-fact structure and reproduces both failures. With evidence that the structure is sufficient to reproduce the failures, we develop a target-misspecification theory that explains both failures from a single deterministic gradient bias amplified by batch size, and use it to motivate a minimal compatibility-based correction, \textit{CorLIP}: text is treated as positive whenever its clinical assertions agree with the image, with no architecture changes. Beyond benchmark improvements, CorLIP largely removes the post-peak degradation, makes large-batch training succeed where it previously collapsed, substantially amplifies the gains from more data, and is the only configuration that remains stable up to \(B{=}2048\). The correction transfers from chest X-ray to 3D CT modalities.

\end{abstract}

\section{Introduction}

Contrastive language--image pretraining (CLIP) has emerged as the dominant paradigm for learning general-purpose visual representations from paired image--text data~\citep{radford2021learning}, and the same contrastive objective is widely used in radiology vision--language pretraining (VLP) across imaging modalities, including X-ray~\citep{zhang2022contrastive,huang2021gloria,tiu2022expert,boecking2022making,chen2024vision,wang2026data}, CT~\citep{hamamci2026generalist,blankemeier2026merlin,zhang2026expert}, MRI~\citep{he2025omnimri} and multi-modality images~\citep{zhao2025towards,agrawal2025pillar}. These systems usually inherit a paired-identity assumption from CLIP: within a batch, the paired text is the only positive for an image, and every other text is treated as a negative. This assumption is approximately correct for web image--caption data, where captions and image content are less likely to overlap, but its validity on radiology image--report data has not been systematically examined. Prior work tends to attribute medical CLIP's difficulties to a broad medical-vs-general domain gap~\citep{lu2025radclip,kumar2024improving,chen2024unicrossadapter}. We argue this gap has a precise and identifiable structure that helps explain why the standard recipe is systematically biased.

We systematically study the training dynamics of medical CLIP and document two failure modes that current contrastive theory does not address. First, the \emph{rise-then-fall dynamic}: zero-shot performance rises, peaks, and then degrades while training loss decreases monotonically, even though a supervised classifier on the identical backbone shows no late-training drop, ruling out generic overfitting. Second, \emph{large-batch collapse}: training fails to learn (AUC near random, loss near $\log B$) at batch sizes that are routinely beneficial for general-domain CLIP~\citep{radford2021learning,zhai2023sigmoid,cherti2023reproducible}. Existing literature has noted pieces of these phenomena without systematic analysis~\citep{wang2022medclip,ko2025bringing,wald2025comprehensive,zhang2026expert} or attributed the rise-then-fall pattern to overfitting~\citep{fisher2025vision}, but neither failure has a unified explanation. Together, these upstream failures cap downstream performance and prevent medical CLIP from leveraging the growing scale of image--report data.

These observations point to a single suspect: the contrastive target. We hypothesize that both failures arise from a structural property of the data rather than from medical-specific appearance, language, or annotation. Specifically, radiology image--report pairs share what we call a \emph{shared-schema fact-bundle structure}: anatomy is the same across images (lungs, heart, mediastinum, pleura), each anatomical region carries a finding state drawn from a small clinical vocabulary (e.g., effusion present or absent, cardiomegaly present or absent), and each report bundles assertions about many such states simultaneously. Under this structure, multiple unpaired reports can match a given image on most asserted findings, so paired identity is systematically biased rather than merely noisy. The hypothesis is falsifiable: any data sharing the same three ingredients should fail in the same way, regardless of domain. To test this, we construct \emph{CLEVR-Radiology}~\citep{johnson2017clevr}, a synthetic testbed that retains the three ingredients but strips every radiology-specific factor such as image appearance, clinical jargon, annotation noise, and scale. Both failures recur identically on CLEVR-Radiology, supporting the hypothesis that the structure, rather than radiology-specific artifacts alone, can drive the failures. We then formalize this as a target-misspecification theory that derives both failures from a single deterministic gradient bias amplified by batch size.

Guided by the theory, we propose \textit{CorLIP}, a minimal correction that replaces CLIP's paired identity with a fact-compatibility target: a text is treated as positive whenever its clinical assertions agree with the image. The correction is compatibility-based rather than similarity-based: a text need not summarize the whole image or be globally similar to the paired report; it only needs to assert facts that are true of the image. This contrasts with prior medical contrastive approaches that mitigate false negatives or improve representations through similarity computations, knowledge integration, or auxiliary alignment~\citep{wu2023medklip,liu2024imitate,liu2023improving,ko2025bringing}. Beyond benchmark improvements on chest X-ray classification and grounding, three insights follow: the rise-then-fall dynamic largely disappears; large-batch training, previously seen as harmful for medical CLIP, succeeds and remains stable up to $B{=}2048$; and additional training data delivers substantial rather than marginal gains. The correction generalizes across radiology modalities, from chest X-ray (CXR) to 3D CT.

\textbf{Contributions.}
\emph{(i) Diagnostic.} We identify two failures of medical CLIP (rise-then-fall, large-batch collapse) and isolate their cause as structural via \textit{CLEVR-Radiology}, a controlled synthetic testbed.
\emph{(ii) Theory.} A target-misspecification theory derives both failures from a single deterministic gradient bias, with two verified predictions: a quantitative loss-floor match and a mechanistic shortcut heatmap drift.
\emph{(iii) Correction.} We propose \textit{CorLIP}\footnote{Project repository: \url{https://github.com/yhygao/CorLIP}}, a minimal compatibility-based correction, which requires no change to architecture or optimizer, and only changes the batch target matrix constructed with an offline fact parser. Beyond benchmark improvements on chest X-ray classification and grounding, CorLIP recovers scaling regimes that do not work with vanilla medical CLIP: large-batch training succeeds where it previously collapsed (stable up to $B{=}2048$), additional data delivers substantial rather than marginal gains, and the correction generalizes from chest X-ray to 3D CT modalities.

\section{Observations}
\label{sec:obs}

We document the training dynamics of vanilla medical CLIP on MIMIC-CXR~\citep{johnson2019mimic} as a representative radiology pretraining corpus. We sweep over batch size (moderate $B{=}128$, large $B{=}1024$) and text granularity (full report, impression section, individual sentence), using a standard recipe (ViT-B + BERT~\citep{dosovitskiy2020image,devlin2019bert} with vanilla symmetric CLIP loss). Zero-shot 14-class mean AUC~\citep{irvin2019chexpert,smit2020combining} measures downstream quality; a supervised multi-label classifier on the same backbone serves as a non-contrastive control. Full setup is in Appendix~\ref{app:exp}.

\begin{figure}[t]
\centering
\includegraphics[width=\linewidth]{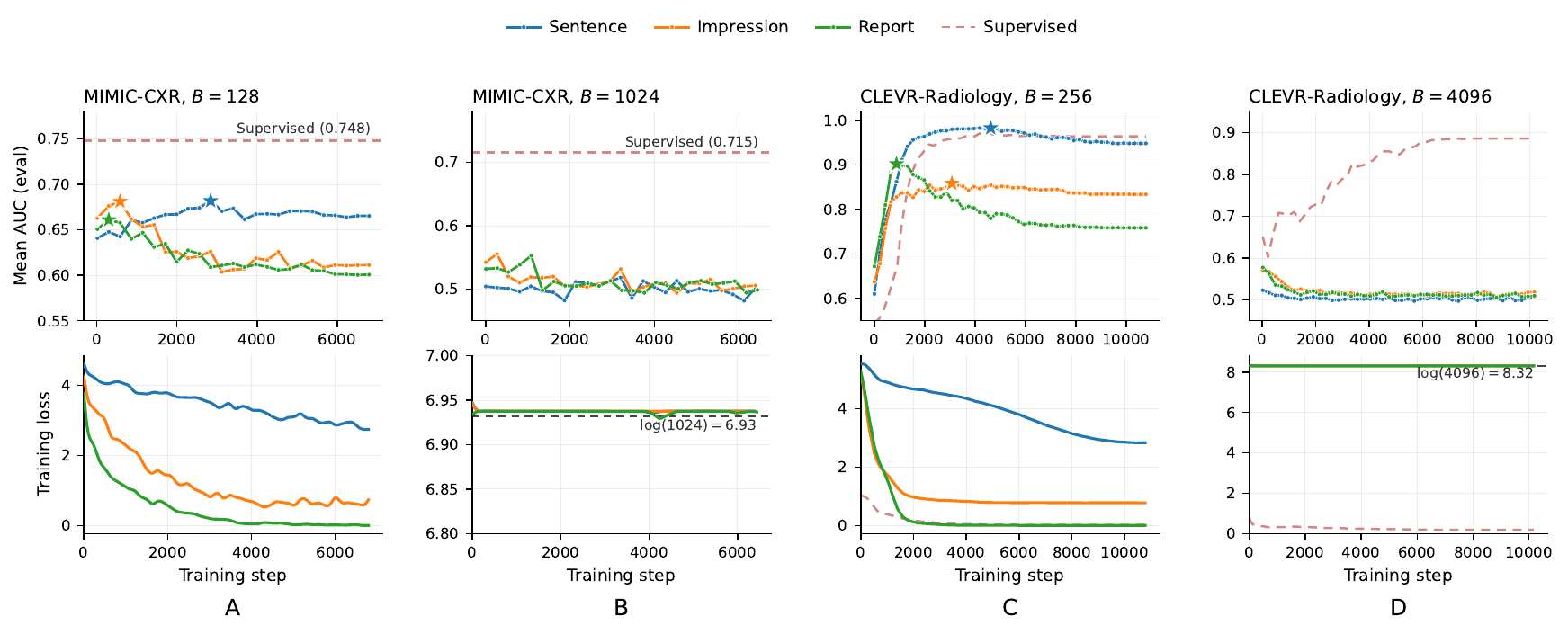}
\caption{\textbf{Two dynamics of vanilla CLIP.} On MIMIC-CXR (A, B) and \textit{CLEVR-Radiology} (C, D), moderate-batch training shows rise-then-fall evaluation while loss keeps decreasing; large-batch training collapses, with AUC near random and loss near \(\log B\). Stars mark peak-AUC checkpoints. The supervised and CLIP losses are not directly comparable in absolute value; we contrast their training trajectories, not their levels.}

\label{fig:dynamics}
\end{figure}

\textbf{Two failure modes.}  At moderate batch ($B{=}128$), zero-shot AUC rises then falls while training loss decreases monotonically; we call this \textbf{O1: rise-then-fall} (Figure~\ref{fig:dynamics}(A)). At large batch ($B{=}1024$), AUC remains near random and loss stays close to $\log B$ throughout, with no learning; we call this \textbf{O2: large-batch collapse} (Figure~\ref{fig:dynamics}(B)). Both patterns hold across all three text granularities.

\textbf{Why these dynamics are puzzling.} Neither failure is explained by ordinary causes. For O1, a supervised classifier on the same data and backbone shows no late-training degradation, and vanilla CLIP's peak stays well below this supervised baseline, arguing against generic overfitting as the sole explanation. For O2, large-batch training is routinely beneficial for general-domain CLIP~\citep{radford2021learning,zhai2023sigmoid,cherti2023reproducible}; here it prevents learning entirely. The contrastive target is therefore the natural suspect.

\textbf{Structural hypothesis.}
What data structure would make the contrastive target fail this way? Unlike web image--caption data, where captions across images rarely overlap, radiology image--report pairs share a \emph{shared-schema fact-bundle structure}: images depict a largely fixed anatomy (lungs, heart, mediastinum, pleura); each region carries finding states from a small clinical vocabulary; and reports bundle assertions about many such states. A short assertion like ``no pleural effusion'' is supported by many images in the batch; a longer report like ``Mild cardiomegaly. No pleural effusion. Lungs are clear.'' makes three independent assertions about distinct anatomical regions, and an unpaired report typically matches the anchor image on most of them. Yet vanilla CLIP treats every off-diagonal text as a pure negative, pushing apart what is in fact compatible. We hypothesize this structural property, not any medical-specific artifact, drives both failures, and that the bias compounds with batch size.

\textbf{\textit{CLEVR-Radiology} isolates the structure.}
To test this without confounders from anatomy variation, acquisition artifacts, clinical language, or incomplete annotation, we designed a controlled experiment: \emph{CLEVR-Radiology} is a synthetic testbed where only the three structural ingredients are preserved by construction. Each image contains a fixed set of simple objects (two cubes, a sphere, a cylinder) playing the role of anatomy; each object carries attribute slots (color, size, material) drawn from small vocabularies with one normal and several abnormal values, playing the role of finding states; a template-based generator emits sentence-, impression-, and report-level texts bundling assertions about these slots. Exact fact metadata is recorded for every (image, text) pair (Appendix~\ref{app:clevr_data}). On CLEVR-Radiology, both failures recur (Figure~\ref{fig:dynamics}(C, D)): at moderate batch, evaluation rises then falls while loss decreases; at large batch ($B{=}4096$), AUC stays near chance with loss near $\log(4096)$. The structure alone is sufficient, supporting the hypothesis. Section~\ref{sec:theory} attributes both failures to a single gradient bias that compounds with batch size.

\section{Theory: A Single Misspecified Target Explains Both Failures}
\label{sec:theory}

Section~\ref{sec:obs} located the cause of O1 and O2 in a structural property of the data. We now make this precise. Under that structure, vanilla CLIP's paired-identity target is misspecified in a way that is \emph{deterministic} (present at every training step) and \emph{batch-amplified} (worsens as $B$ grows). A single decomposition of the vanilla CLIP gradient recovers O1 and O2 as its two regimes.

\subsection{Setup: Fact Compatibility}
\label{sec:theory:setup}

Vanilla CLIP treats each image as having exactly one matching text. In radiology, many texts can be semantically true of the same image; we formalize this and use it to define an alternative reference target alongside vanilla's paired identity.

\textbf{Image-text setup.}
A training batch contains $B$ image--text pairs $\{(x_i, t_i)\}_{i=1}^B$. Let $s_{ij}$ be the image--text similarity logit, $p_i^\theta(j) = \exp(s_{ij}) / \sum_\ell \exp(s_{i\ell})$ the row softmax under model parameters $\theta$, and $\ell_i(q_i) = -\sum_j q_i(j)\log p_i^\theta(j)$ the row cross-entropy against any target distribution $q_i$. Vanilla CLIP uses the paired-identity target $e_i(j) = \mathbf{1}\{j=i\}$. We analyze image-to-text rows; text-to-image is symmetric.

\textbf{Compatibility kernel.}
Each image $x_i$ has a latent fact state $z_i = (z_{i,1},\ldots,z_{i,F})$ over $F$ discrete attribute slots, and each text $t$ asserts a set $A(t)$ of (slot, value) pairs. We say $x_i$ \emph{entails} $t_j$, written $z_i \models A(t_j)$, if every $(a,v) \in A(t_j)$ satisfies $z_{i,a} = v$. In radiology, a text's meaning reduces to its asserted facts: two texts that assert the same facts are semantically equivalent regardless of surface form (e.g., ``cardiomegaly'' and ``enlarged heart'' mean the same for any image), and an image semantically supports a text exactly when the image's facts satisfy the text's assertions. The \emph{compatibility kernel} $K_{ij} = \mathbf{1}\{z_i \models A(t_j)\}$ formalizes this as the simplest fact-level equivalence, collapsing all surface variation. With $m_i = \sum_{j=1}^B K_{ij}$ compatible texts in a batch of $B$, the row-normalized \emph{compatibility reference} $q_i^K(j) = K_{ij}/m_i$ is the simplest semantic target consistent with this kernel: it asks the model to assign equal probability $1/m_i$ to every text in the batch that is semantically supported by the image, and zero to incompatible texts, treating semantically equivalent texts as interchangeable. Vanilla CLIP's $e_i$ implicitly assumes $m_i = 1$; under the structure of Section~\ref{sec:obs}, $m_i > 1$ is typical.

\textbf{Bundled reports.}
For multi-finding texts, exact entailment is too strict: a report may share most asserted facts with the anchor while disagreeing on one. We therefore also define a soft reference $q_i^R$ via the partial-agreement fraction $R_{ij} = |A(t_j)|^{-1}\sum_{(a,v)\in A(t_j)} \mathbf{1}\{z_{i,a}=v\}$, and write $q_i^\star \in \{q_i^K, q_i^R\}$ when arguments apply to either (Appendix~\ref{app:report_density}). The analysis below uses $q_i^K$ for clarity.

\textbf{Notation across domains.}
\emph{In radiology}, slots index findings such as cardiomegaly and pleural effusion with values in \{present, absent, uncertain\}; ``no pleural effusion'' asserts a single (slot, value) pair, so any effusion-free image entails it. \emph{In CLEVR-Radiology}, slots index object attributes (color, size, material) with small per-attribute vocabularies, and $z_i$ is fully observed. \emph{In general-domain CLIP}, captions are largely instance-specific, so $A(t_j)$ is essentially unique per image, $m_i \approx 1$.

\subsection{Decomposing Vanilla CLIP's Gradient}
\label{sec:theory:bias}

In logit space, vanilla CLIP's gradient can be decomposed into two parts: a \emph{semantic} part that drives the model to score equally any text that is true of the image, and a fixed \emph{identity} part that rewards singling out the paired text from those compatible alternatives.

\begin{lemma}[Vanilla CLIP's gradient decomposition]
\label{lem:gradient_bias}
For any compatibility reference $q_i^\star$, vanilla CLIP's logit-space gradient descent direction decomposes as
\[
    -\nabla_{s_i}\ell_i(e_i)
    \;=\;
    \underbrace{e_i - p_i^\theta}_{d_i^{\mathrm{van}}(\theta)}
    \;=\;
    \underbrace{(q_i^\star - p_i^\theta)}_{d_i^{\mathrm{sem}}(\theta)}
    \;+\;
    \underbrace{(e_i - q_i^\star)}_{d_i^{\mathrm{id}}}.
\]
\end{lemma}

The lemma follows from $\nabla_{s_i}\ell_i(e_i) = p_i^\theta - e_i$ by adding and subtracting $q_i^\star$ (Appendix~\ref{app:proof_lemma_gradient}). The decomposition isolates a structural fact: $d_i^{\mathrm{id}}$ depends only on the batch, so vanilla CLIP carries a deterministic gradient bias relative to the compatibility reference, independent of training state.

The two components play different roles. $d_i^{\mathrm{sem}}(\theta)$ is the \emph{semantic learning signal}: it is the descent direction of training directly against $q_i^K$, and vanishes during semantic learning, as the model approaches scoring all $m_i$ compatible texts equally. $d_i^{\mathrm{id}}$ is what remains: it pushes up the paired logit $s_{ii}$ and pushes down every equally compatible but unpaired $s_{ij}$. This is a wrong learning signal, asking the model to distinguish semantically equivalent texts via non-fact-level features. The next two subsections show how $d_i^{\mathrm{sem}}$ and $d_i^{\mathrm{id}}$ produce O1 and O2.

\subsection{Semantic Floor and Shortcut Incentives}
\label{sec:theory:o1}

Once vanilla CLIP has saturated semantic learning, the only remaining gradient rewards non-semantic features that single out the paired text. This explains O1.

An ideal fact-level encoder would assign equal similarity to all compatible texts, since they are semantically equivalent for the image. We call logits with this property \emph{purely semantic}: $K_{ij}=K_{ik}=1 \Rightarrow s_{ij}=s_{ik}$. Under this ideal, $p_i^\theta(i) \leq 1/m_i$ (Appendix~\ref{app:proof_prop2}), with equality when $p_i^\theta = q_i^K$ (incompatible logits $\to -\infty$). The vanilla loss is therefore bounded below by $\log m_i = H(q_i^K)$, the entropy of the compatibility reference; we call this the \emph{semantic floor}. As the model approaches the floor, $p_i^\theta = q_i^K$, so by Lemma~\ref{lem:gradient_bias} the vanilla CLIP's negative gradient reduces to $d_i^{\mathrm{id}}$: the semantic part vanishes and only the fixed identity residual remains.

\begin{proposition}[Shortcut reward at the semantic floor]
\label{prop:semantic_floor}
At $p_i^\theta = q_i^K$, for any feature direction $u$ that raises only the paired logit $s_{ii}$,
    $\frac{\partial \ell_i(e_i)}{\partial u} \;=\; -\!\left(1 - \frac{1}{m_i}\right), \qquad \frac{\partial \ell_i(q_i^K)}{\partial u} \;=\; 0$.
\end{proposition}

Proof in Appendix~\ref{app:proof_prop2}. In the encoder, such $u$ correspond to text-side cues (specific wording, sentence ordering) or image-side cues (border markers, medical devices, in-image text), not the underlying clinical facts. Below the semantic floor, the symmetric vanilla loss forces both encoders to break fact-level clustering: $x_i$ must rank $t_i$ above other compatible $t_j$, so the text encoder scatters semantic-equivalent texts along wording axes; $t_i$ must rank $x_i$ above other images that entail it, so the image encoder scatters same-finding images along image-specific cues. This is not benign overfitting: zero-shot prompts carry none of these training-side cues, so encoder capacity that should encode \emph{what} fact an image asserts is reallocated to encoding \emph{which} text it was paired with, and clean class prompts no longer find their image neighbors. The longer training stays below the floor, the wider this misalignment grows, producing the rise-then-fall.

\textbf{Predicting O1.}
Before the floor, $d_i^{\mathrm{sem}}$ dominates the gradient and improves both loss and zero-shot evaluation; the fixed $d_i^{\mathrm{id}}$ is present throughout but overshadowed. When approaching the semantic floor, the semantic component vanishes and vanilla training shifts to rewarding these shortcut cues, so zero-shot classification can decline while training loss keeps falling. The persistent $d_i^{\mathrm{id}}$ also caps the best vanilla checkpoint: optimization effort is constantly diverted to shortcuts even when $d_i^{\mathrm{sem}}$ dominates, leaving vanilla's peak below the supervised baseline. The supervised classifier has no analogue of $d_i^{\mathrm{id}}$ (its multi-label target carries no paired-instance direction), so its peak is not similarly capped, and it shows no late-training drift.

\begin{figure}[t]
\centering
\begin{minipage}[b]{0.26\linewidth}\centering
  \includegraphics[width=\linewidth]{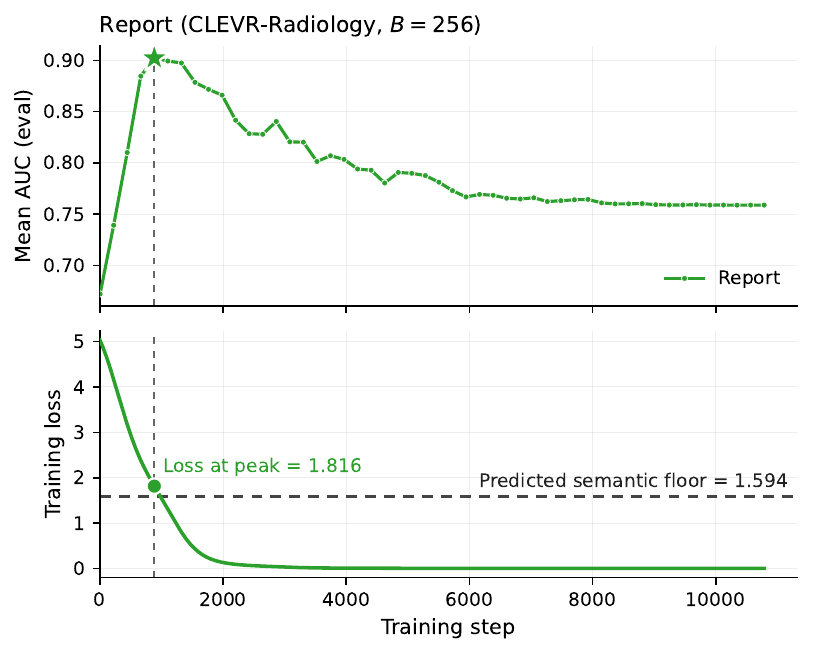}\\
  {\small (A) Report semantic floor}
\end{minipage}\hfill
\begin{minipage}[b]{0.26\linewidth}\centering
  \includegraphics[width=\linewidth]{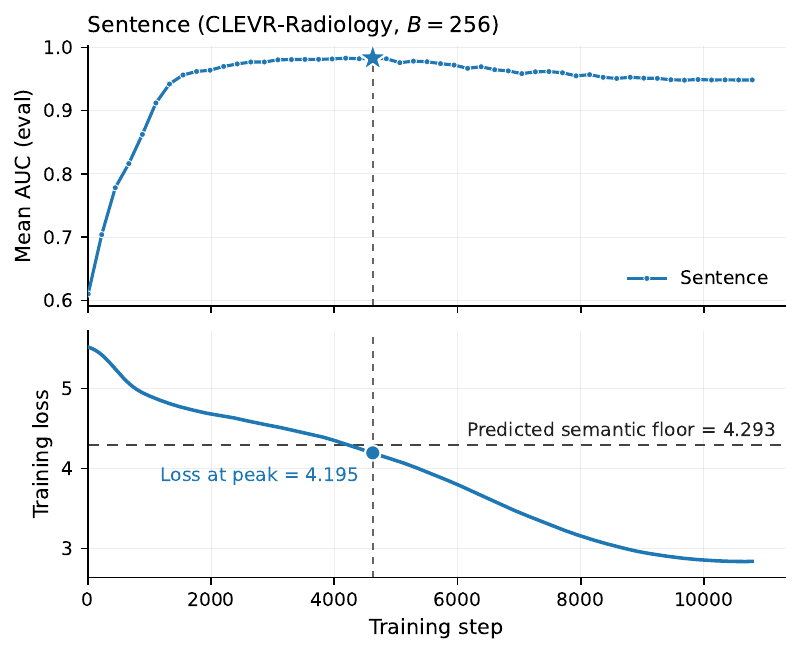}\\
  {\small (B) Sentence semantic floor}
\end{minipage}\hfill
\begin{minipage}[b]{0.45\linewidth}\centering
  \includegraphics[width=\linewidth]{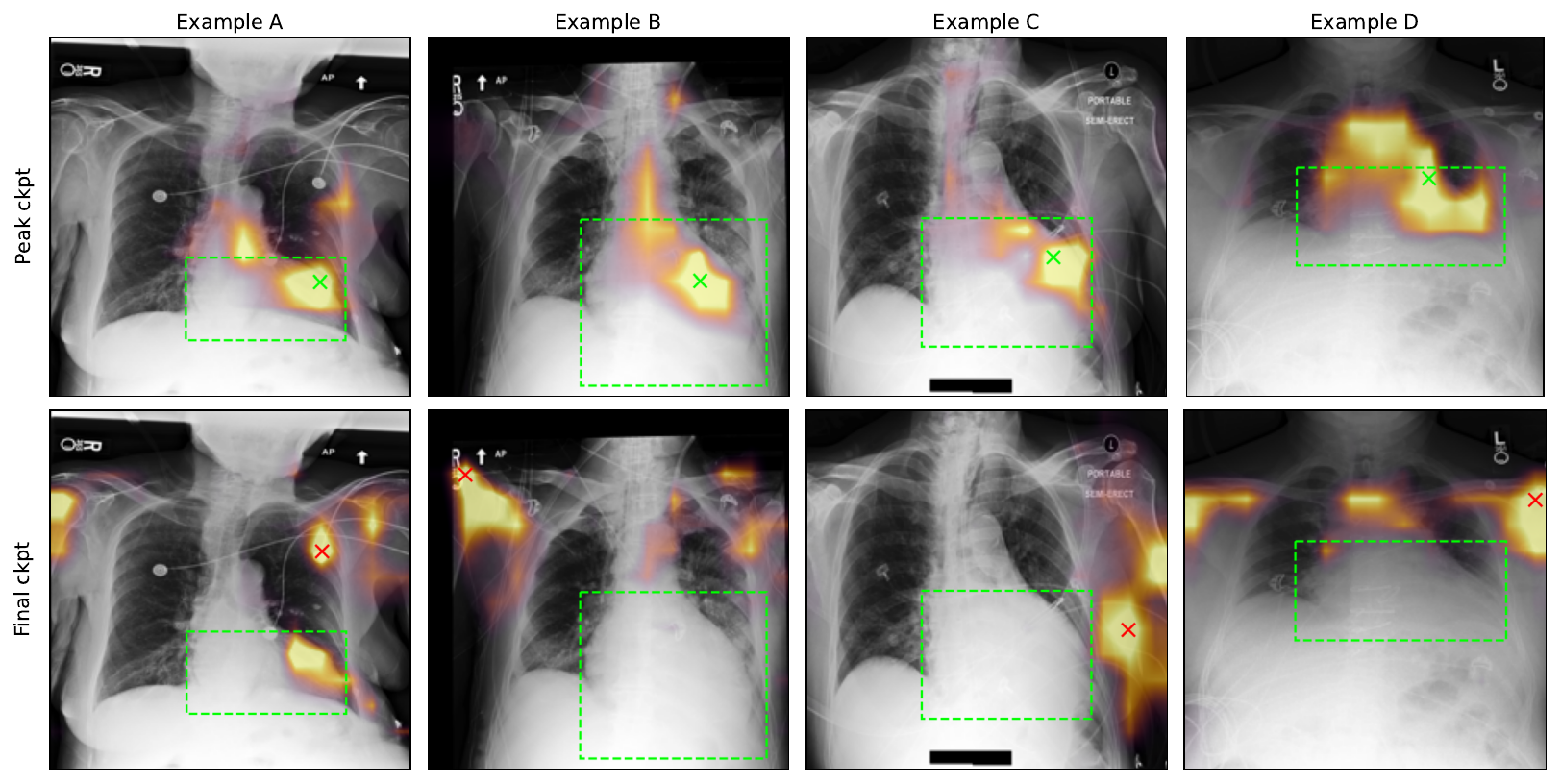}\\
  {\small (C) Shortcut drift}
\end{minipage}
\caption{\textbf{Evidence for the semantic-floor and shortcut-learning predictions.} (A, B) On \textit{CLEVR-Radiology}, the eval-peak loss aligns with the generator-computed semantic floor. (C) On MIMIC-CXR, cardiomegaly grounding shifts from the clinical region at peak to irrelevant cues at final.}
\label{fig:theory_evidence}
\end{figure}

\textbf{Evidence.}
Two complementary tests confirm the theory. On CLEVR-Radiology, the theory predicts the floor $\log m_i$ from generator metadata before any training, and predicts that loss below this floor coincides with evaluation decline. Real training dynamics confirms this (Figure~\ref{fig:theory_evidence}(A, B)). On CXR data, the latent state is hidden, but the predicted shortcuts are visible directly in the model's similarity map: phrase-grounding heatmaps after the peak show attention drifting from the cardiac silhouette to markers and devices, with cardiomegaly pointing-game accuracy dropping from 0.394 at peak to 0.213 at final (a 46\% relative loss) while training loss continues to decrease (Figure~\ref{fig:theory_evidence}(C)).

\subsection{Batch-amplified Identity Pressure}
\label{sec:theory:o2}

At large batch, the identity component dominates the gradient from the very first step; the model can fail to build semantic representations. This explains O2.

At initialization, $p_i^\theta \approx \pi_B$, the uniform distribution over the $B$ batch texts, so Lemma~\ref{lem:gradient_bias} specializes to
\[
    e_i - \pi_B \;=\; \underbrace{(q_i^K - \pi_B)}_{d_i^{\mathrm{sem}}(0)} \;+\; \underbrace{(e_i - q_i^K)}_{d_i^{\mathrm{id}}},
\]
and the two components are now exactly orthogonal (Appendix~\ref{app:proof_prop3}). $d_i^{\mathrm{sem}}(0)$ raises every compatible text and lowers every incompatible one (the semantic direction that the model would learn); orthogonality lets us cleanly compare component magnitudes.

\begin{proposition}[Batch-amplified identity pressure]
\label{prop:batch_pressure}
Let $\omega$ be the probability that an unpaired text is compatible with an independent image, so $\mathbb{E}[m_i] \sim B\omega$. At initialization,
\[
    \Gamma_i(0) \;=\; \frac{\|d_i^{\mathrm{id}}\|_2^2}{\|d_i^{\mathrm{sem}}(0)\|_2^2}
    \;=\; \frac{B(m_i-1)}{B-m_i}
    \;\sim\; \frac{\omega}{1-\omega}\,B.
\]
The semantic share of the vanilla gradient energy, $1/(1+\Gamma_i(0))$, shrinks as $O(1/B)$.
\end{proposition}

Proof in Appendix~\ref{app:proof_prop3}. The fixed $d_i^{\mathrm{id}}$ asks the model to pick one paired text from $m_i - 1$ equally compatible alternatives before any fact-level representation has formed, an ill-posed objective at this stage. The component that could build such a representation has only $O(1/B)$ energy share, so training stalls near the uniform-softmax regime with loss near $\log B$ and AUC near random. Radiology reports usually state each finding as present or absent (e.g., ``cardiomegaly'' vs.\ ``no cardiomegaly''), so any image shares many compatible assertions with many unpaired reports, regardless of finding prevalence; this keeps $\omega$ often far from $0$ and produces severe dilution at large batches. For general-domain CLIP, $\omega \approx 0$ makes $d_i^{\mathrm{id}}$ vanish and the gradient stays essentially semantic, which is why large-batch vanilla CLIP works on web data.

\section{Mitigation: Approximating the Compatibility Reference}
\label{sec:method}

O1 and O2 share one root: the fixed identity residual $d_i^{\mathrm{id}}$. CorLIP shrinks it by replacing the vanilla target $e_i$ with a closer approximation of $q_i^K$, the only modification to the contrastive objective; architecture, optimizer, and batch construction are unchanged.

\textbf{Estimating image states and text assertions.}
The references $q_i^K$ and $q_i^R$ require known fact states $z_i$, which is observed in \textit{CLEVR-Radiology} but latent in real radiology data. The paired report, however, is itself the radiologist's reading of $z_i$ from the image, so CorLIP recovers $\hat z_i = E_{\mathrm{state}}(r_i)$ and $\hat A(t_j) = E_{\mathrm{text}}(t_j)$ by parsing reports offline as a preprocessing step. The construction pipeline can be extended to other radiology modalities by swapping the extractor (Figure~\ref{fig:app_method_visualization}).

\textbf{Choosing fact granularity.} When are facts from two reports compatible? The answer is non-trivial and depends on the chosen granularity. At coarse granularity, ``left pleural effusion'' and ``right pleural effusion'' can both map to ``pleural effusion'' and be treated as compatible; this expands the compatible pool but also asserts equivalence between clinically distinct sub-findings, which may not match how the downstream task evaluates them. At fine granularity, these distinctions can be preserved, but the compatible pool shrinks and the shared structure across related sub-findings is no longer reflected in the target. No single choice is optimal across all evaluations.

For CXR we instantiate CorLIP with two practical schemas spanning this tradeoff. \textit{CheXbert-13d}~\citep{smit2020combining} is a coarse 13-label schema aligned with CheXpert-style evaluations; we drop the aggregation ``No Finding'' label since it is not an independent finding. \textit{MedGemma-30d} is a finer schema we define using MedGemma-27B-Text-IT~\citep{sellergren2025medgemma} as a structured text parser, covering more findings and distinguishing laterality. Both schemas operate on a flat list of present/absent finding facts; we deliberately leave richer compatibility kernels (severity grading, anatomy/finding hierarchies, inter-finding dependencies) to future work to isolate the contribution of target correction itself. Unless stated otherwise, corrected CXR results use MedGemma-30d as the less benchmark-coupled default.

\textbf{Scoring compatibility.}
Given a schema, we score only facts explicitly asserted by the candidate text; unmentioned findings are not treated as contradictions. For sentence-level texts, compatibility is close to binary since a sentence asserts one or two facts: \(\hat C_{ij}^{\mathrm{hard}} = \mathbf{1}\{\hat z_i \models \hat A(t_j)\}\) approximates \(K\), so multiple sentences supporting the same finding share target mass instead of competing as negatives. For impressions and reports, exact entailment is too strict because a bundled text may agree on most facts while disagreeing on one, so partial agreement \(\hat C_{ij}^{\mathrm{soft}} = |\hat A(t_j)|^{-1}\sum_{(a,v)\in \hat A(t_j)} \mathbf{1}\{\hat z_{i,a}=v\}\) approximates \(R\); a report sharing four of five asserted findings then receives partial target mass rather than being a pure negative.

\textbf{Contrastive targets.}
Given a nonnegative \(\hat C\), we row- and column-normalize to form targets \(\hat q^{I\to T}_{ij} = \hat C_{ij}/\sum_\ell \hat C_{i\ell}\) and \(\hat q^{T\to I}_{ji} = \hat C_{ij}/\sum_\ell \hat C_{\ell j}\) that drop into the standard symmetric CLIP cross-entropy. We retain the paired diagonal; handling of empty or non-clinical texts, uncertainty, unmentioned findings, and extractor conventions is in Appendix~\ref{app:compat_edges}.

\textbf{Why sentence-level correction is best aligned.}
The residual under target \(\hat q_i\) is \(\hat q_i - q_i^\star\) (Lemma~\ref{lem:gradient_bias}); the closer \(\hat q_i\) is to the reference, the smaller this residual. Sentence-level texts are closest to atomic assertions, so the compatibility decision is least ambiguous and hard entailment is closest to \(K\). Bundled texts mix findings, omissions, uncertainty, and stylistic conventions, forcing many fact-level agreements into one scalar. By Proposition~\ref{prop:batch_pressure}, this residual is further amplified by batch size, so sentence-level CorLIP is predicted to be the most stable configuration as \(B\) grows.

\section{Experiments}
\label{sec:exp}

\textbf{Setup.}
\textit{Data and training.} We train on $60\text{k}$ frontal MIMIC-CXR image--report pairs~\citep{johnson2019mimic}. Report uses FINDINGS+IMPRESSION, impression uses IMPRESSION only, and sentence samples one report sentence per iteration. \textit{Controlled comparison.} To isolate the contribution of the contrastive target, all methods share the same DINOv3-initialized ViT-B image encoder~\citep{simeoni2025dinov3}, BioClinicalBERT text encoder~\citep{alsentzer2019publicly}, training data, augmentations, optimizer, and schedule. Methods differ \emph{only} in their contrastive target. Classical CXR VLP objectives are absorbed into this controlled pipeline rather than compared as separate checkpoints: CheXzero~\citep{tiu2022expert} and ConVIRT~\citep{zhang2022contrastive} training reduce to vanilla CLIP at report/impression and sentence granularity respectively, so we reinstantiate their paired-identity objective under our pipeline rather than reusing original checkpoints, which differ in training recipe. \textit{Evaluation.} For zero-shot classification, we test in-domain on MIMIC-CXR 14-class and out-of-distribution on CheXpert 14-class~\citep{irvin2019chexpert} and VinDr 28-class~\citep{nguyen2022vindr}. For dense evaluation, we use phrase grounding on MS-CXR bounding boxes~\citep{boecking2022making}. Full hyperparameters are in Appendix~\ref{app:exp}.

The experiments are designed as theory checks, not architecture search. They ask whether correcting the target changes the training trajectory, whether data scale helps once the target is corrected, and whether large batches become useful when compatible texts are no longer repelled.

\begin{figure}[t]
\centering
\begin{minipage}[t]{0.52\linewidth}
\centering
\vspace{0pt}
\includegraphics[width=\linewidth]{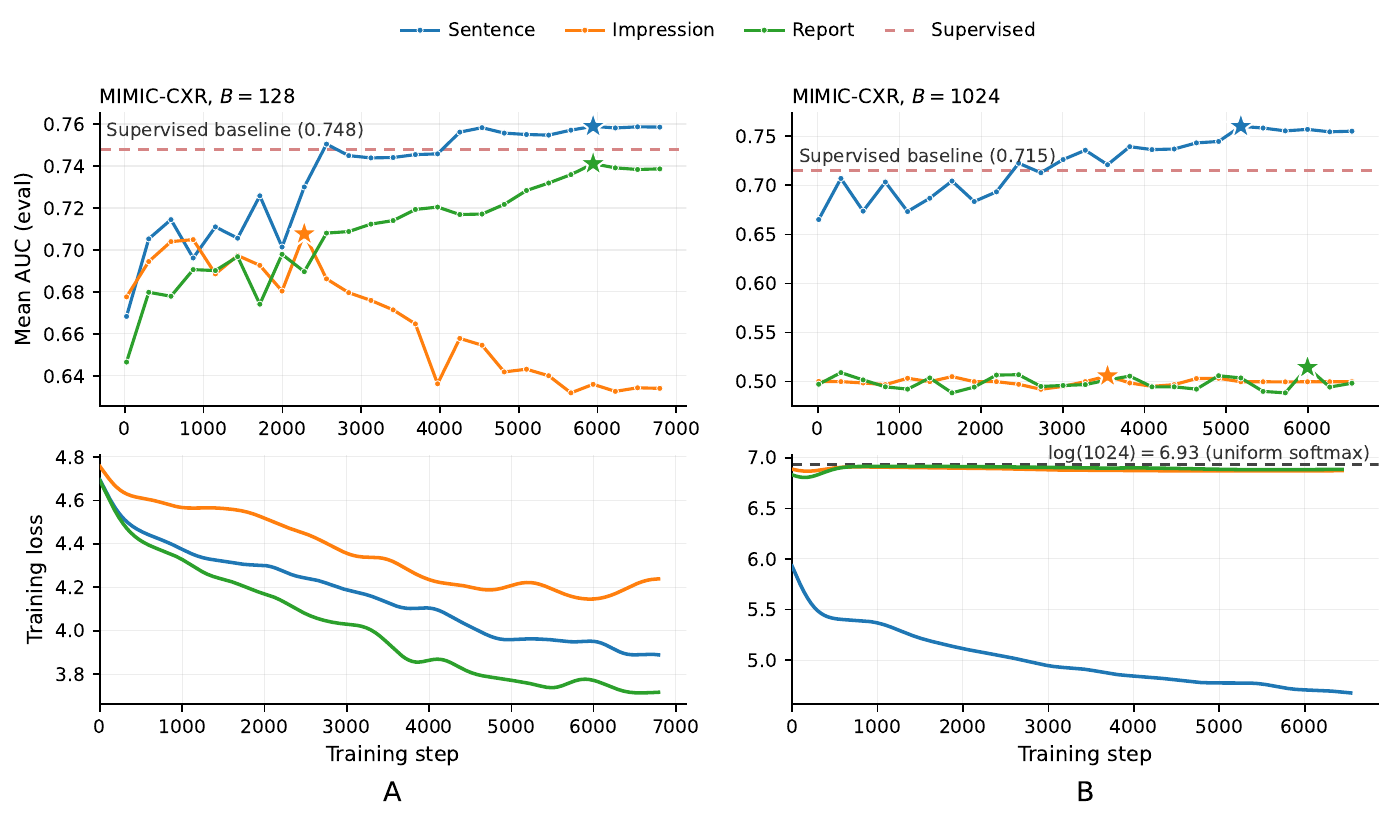}
\captionof{figure}{\textbf{Corrected CLIP training dynamics on MIMIC-CXR.} (A) CorLIP at \(B{=}128\). Sentence- and report-level rise-then-fall are largely fixed. (B) CorLIP at \(B{=}1024\). Sentence CorLIP trains successfully at large batch size. Corrected with CorLIP-MG.}
\label{fig:corrected_dynamics}
\end{minipage}\hfill
\begin{minipage}[t]{0.45\linewidth}
\centering
\vspace{0pt}
\tiny
\setlength{\tabcolsep}{2.6pt}
\renewcommand{\arraystretch}{1.05}
\begin{tabular}{@{}l l ccc ccc@{}}
\toprule
& & \multicolumn{3}{c}{MIMIC AUC} & \multicolumn{3}{c}{MIMIC F1} \\
\cmidrule(lr){3-5} \cmidrule(lr){6-8}
Method & Text & pk & fn & $\Delta$ & pk & fn & $\Delta$ \\
\midrule
Vanilla CLIP            & Rep.     & 66.1          & 60.1          & $-$6.0  & 28.9          & 23.6          & $-$5.4 \\
Vanilla CLIP            & Imp. & 68.1          & 61.1          & $-$7.0  & 28.7          & 23.9          & $-$4.9 \\
Vanilla CLIP            & Sen.   & 68.2          & 66.5          & $-$1.7  & 28.5          & 27.5          & $-$0.9 \\
\midrule
SAT~\citep{liu2023improving}          & Rep.   & 67.9          & 59.3          & $-$8.7  & 29.9          & 24.2          & $-$5.7 \\
MedCLIP~\citep{wang2022medclip} & Sen. & 77.4          & 77.2          & $-$0.2  & 44.7          & 45.0          & $+$0.3 \\
Ko\&Park~\citep{ko2025bringing}    & Rep.   & 78.9          & \underline{77.9} & $-$1.0  & \underline{45.4} & \underline{46.8} & $+$1.3 \\
\midrule
CorLIP-CB          & Rep.     & 77.3          & 75.4          & $-$1.9  & 38.6          & 42.8          & $+$4.2 \\
CorLIP-MG          & Rep.     & 74.1          & 73.9          & $-$0.2  & 33.0          & 32.5          & $-$0.5 \\
CorLIP-CB          & Imp. & \underline{79.0} & 72.5          & $-$6.5  & 42.1          & 42.2          & $+$0.0 \\
CorLIP-MG          & Imp. & 70.8          & 63.4          & $-$7.4  & 32.7          & 27.5          & $-$5.2 \\
CorLIP-CB          & Sen.   & \textbf{81.6} & \textbf{81.4} & $-$0.2  & \textbf{48.5} & \textbf{48.1} & $-$0.4 \\
CorLIP-MG          & Sen.   & 75.8          & 75.8          & 0.0     & 35.5          & 35.5          & 0.0 \\
\bottomrule
\end{tabular}
\par\vspace{0.4em}
\captionof{table}{\textbf{In-domain MIMIC-CXR comparison.} Peak (pk), final (fn), and delta are shown. The pk and fn checkpoints are selected by AUC; F1 is reported at the same checkpoint.}
\label{tab:bs128_indomain}
\end{minipage}
\end{figure}

\textbf{Correcting the target changes the trajectory, not only the score.}
At $B{=}128$, the key change is trajectory-level. Figure~\ref{fig:corrected_dynamics} shows that sentence-level CorLIP largely removes the post-peak drop seen in vanilla CLIP training (Figure~\ref{fig:dynamics}). Table~\ref{tab:bs128_indomain} quantifies the same effect: vanilla CLIP loses $1.7$--$7.0$ AUC points by the final checkpoint, whereas CorLIP-CB sentence reaches the best peak and final scores on both MIMIC AUC and F1, with only $0.2$ AUC and $0.4$ F1 drop. Report-level correction also reduces vanilla report degradation; impression-level correction is less reliable, suggesting that compatibility quality is not monotonic in text length. Impressions are lossy summaries, neither atomic like sentences nor information-rich enough for stable soft report-level correction. Notably, corrected sentence training exceeds the supervised baseline, showing that the contrastive objective recovers its expected advantage once the target is corrected.

\textbf{Data scaling helps more when it scales the right target.}
Appendix~\ref{app:data_scaling} shows that doubling data from \(60\mathrm{K}\) to \(120\mathrm{K}\) raises vanilla CLIP's best peak AUC only from \(68.2\) to \(70.6\), and all granularities still degrade after the peak. With corrected targets, \(60\mathrm{K}\) CorLIP-CB sentence already reaches \(81.6/81.4\) peak/final AUC, over 10 points above the best \(120\mathrm{K}\) vanilla checkpoint; at \(120\mathrm{K}\), it further improves to \(86.5/86.0\). Data scaling is powerful when it scales the right target, but inefficient when it scales a misspecified one.

\begin{figure}[t]
\centering
\begin{minipage}[t]{0.39\linewidth}
\centering
\vspace{0pt}
\tiny
\setlength{\tabcolsep}{2.8pt}
\renewcommand{\arraystretch}{1.0}
\resizebox{\linewidth}{!}{%
\begin{tabular}{@{}l l ccc@{}}
\toprule
Method & Text & CheXpert & VinDr & MS-CXR PG \\
\midrule
Vanilla CLIP                  & Rep.     & 79.8 & 83.0 & 22.4 \\
Vanilla CLIP                  & Imp.     & 78.7 & 79.6 & 30.8 \\
Vanilla CLIP                  & Sen.     & 82.1 & 83.1 & 30.7 \\
\midrule
SAT~\citep{liu2023improving}                & Rep. & 80.7 & 83.9 & 34.0 \\
MedCLIP~\citep{wang2022medclip}       & Sen. & 80.9 & 82.8 & 32.3 \\
Ko\&Park~\citep{ko2025bringing}         & Rep. & 83.5 & 84.1 & 35.8 \\
\midrule
CorLIP-CB                     & Rep.     & 84.5 & 83.7 & 34.5 \\
CorLIP-MG                     & Rep.     & \underline{86.2} & \textbf{87.3} & \underline{36.9} \\
CorLIP-CB                     & Imp.     & 81.7 & 82.7 & 33.1 \\
CorLIP-MG                     & Imp.     & 85.9 & \underline{85.2} & 34.1 \\
CorLIP-CB                     & Sen.     & 85.6 & 83.7 & 37.7 \\
CorLIP-MG                     & Sen.     & \textbf{86.6} & \textbf{87.3} & \textbf{41.6} \\
\bottomrule
\end{tabular}
}
\par\vspace{0.4em}
\captionof{table}{\textbf{OOD classification and grounding.} Mean AUC for CheXpert and VinDr, mean pointing-game accuracy for MS-CXR phrase grounding.}
\label{tab:bs128_ood}
\end{minipage}\hfill
\begin{minipage}[t]{0.58\linewidth}
\centering
\vspace{0pt}
\begin{minipage}[t]{0.42\linewidth}
\centering
\includegraphics[width=\linewidth]{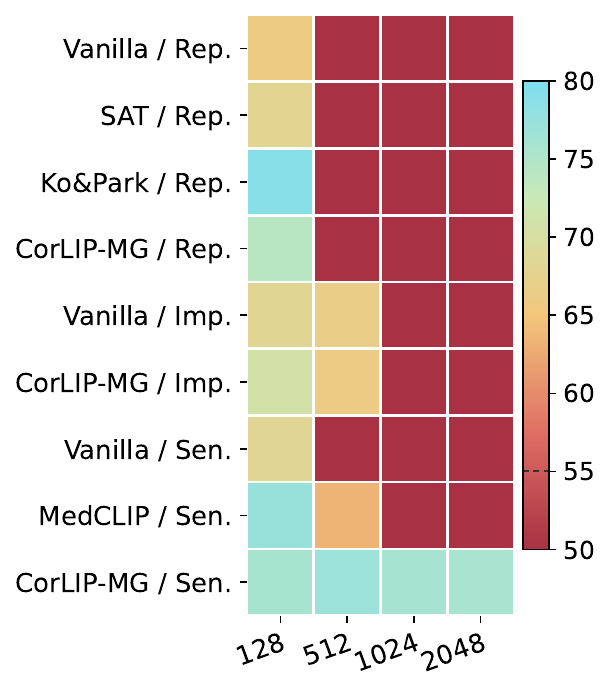}\\
{\small (A) Batch trainability}
\end{minipage}\hfill
\begin{minipage}[t]{0.56\linewidth}
\centering
\includegraphics[width=\linewidth]{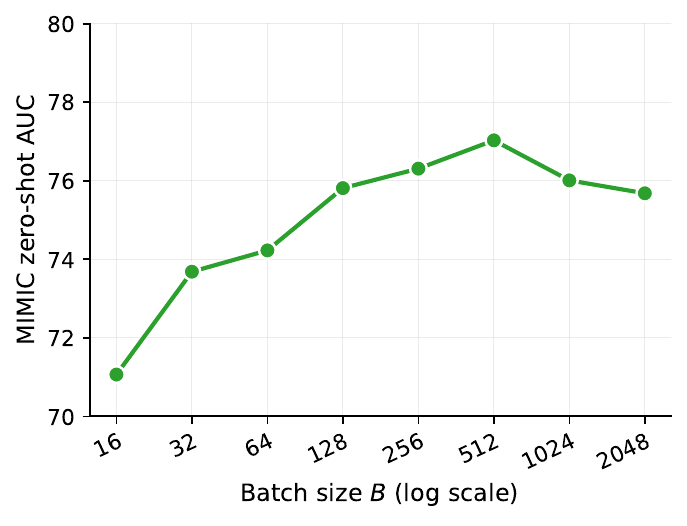}\\
{\small (B) Batch scaling}
\end{minipage}
\par\vspace{0.4em}
\captionof{figure}{\textbf{Large-batch trainability and scaling.} (A) Which methods keep learning or collapse as \(B\) grows. (B) Batch scaling within CorLIP-MG sentence training, the only stable large-batch regime.}
\label{fig:batch_scaling}
\end{minipage}
\end{figure}

\textbf{The gains transfer to out-of-distribution data and dense tasks.}
Table~\ref{tab:bs128_ood} asks whether the stronger MIMIC representation transfers and how it works on dense tasks. CorLIP results in the best result on all evaluated settings: CorLIP-MG sentence on CheXpert, CorLIP-MG report/sentence on VinDr, and CorLIP-MG sentence on MS-CXR grounding. Compared with Table~\ref{tab:bs128_indomain}, CheXbert-based correction is strongest on in-domain MIMIC, whose evaluation labels derive from the same CheXpert-style labeling as its correction schema. MedGemma-based correction is stronger on OOD settings and dense tasks in our runs, making it the stronger fixed default of the two; since the schemas differ in both granularity and extractor, we do not attribute this gap to granularity alone.

\textbf{Large batches become useful once the target is compatible.}
The right column of Figure~\ref{fig:corrected_dynamics} shows the large-batch regime. At $B{=}1024$, report- and impression-level correction still fail to train reliably, while sentence-level correction escapes the uniform-softmax floor and reaches a useful AUC score. Replacing paired identity is therefore not enough if the text unit still bundles many facts into a lossy partial-compatibility score.

Figure~\ref{fig:batch_scaling}(A) broadens the comparison across methods and batch sizes. A few non-sentence configurations still train at $B{=}512$, including MedCLIP and impression-level runs, but they collapse by $B{=}1024$. Among the evaluated configurations, only CorLIP sentence remains stable through $B{=}2048$, where it still achieves competitive peak AUC values. We stop at $B{=}2048$ because larger batches exceed our compute budget.

Figure~\ref{fig:batch_scaling}(B) isolates the stable regime. With corrected sentence targets, larger batches are helpful up to $B{=}512$: peak AUC rises from $71.1$ at $B{=}16$ to $77.0$ at $B{=}512$, then remains stable at $76.0$ and $75.7$ for $B{=}1024$ and $B{=}2048$. The usual lesson from medical CLIP would be to avoid large batches~\citep{zhang2022contrastive,wang2022medclip,ko2025bringing,wald2025comprehensive}, and commonly used CXR batch sizes are often below 128~\citep{tiu2022expert,wang2026data}. Our result changes the diagnosis: large batches are harmful under a misspecified target, but more negatives in large batches become useful again after target correction. The peak arrives much earlier than in web-scale CLIP systems~\citep{radford2021learning,zhai2023sigmoid,cherti2023reproducible}, plausibly because CXR has a smaller effective semantic space, so a few hundred images already provide enough genuinely distinct negatives.

\textbf{Beyond softmax CLIP.} Appendix~\ref{app:siglip} shows the same conclusion for SigLIP-style sigmoid contrastive training: vanilla SigLIP exhibits the same rise-then-fall and large-batch collapse, while our correction can largely mitigate the issue.

\begin{wraptable}{r}{0.38\linewidth}
\centering
\caption{\textbf{3D CT transfer (Merlin).} Peak mean AUC over 30 findings, $\times100$, with 95\% CI.}
\label{tab:ct}
\small
\setlength{\tabcolsep}{10pt}
\begin{tabular}{@{}llc@{}}
\toprule
Text & Method & AUC \\
\midrule
Rep. & Vanilla & 67.9 {\scriptsize(66.4--69.5)} \\
Sen. & Vanilla & 75.8 {\scriptsize(74.5--77.2)} \\
Sen. & CorLIP  & \textbf{79.3} {\scriptsize(78.0--80.6)} \\
\bottomrule
\end{tabular}
\end{wraptable}
\textbf{Correction transfers to 3D CT.}
The bias we identified is target-level, not modality-level. To check this, we apply CorLIP to Merlin 3D CT image--report pretraining with a CT-adapted MedGemma extractor; encoder, preprocessing, optimizer, and schedule are fixed at Merlin's defaults (Appendix~\ref{app:exp_ct}). Report-level vanilla CLIP shows the same rise-then-fall dynamic as on CXR: mean AUC peaks at 67.9 and then degrades by about 5 points while the training loss keeps decreasing. Sentence-level training raises the peak to 75.8, and sentence-level CorLIP reaches 79.3 (Table~\ref{tab:ct}). The CXR pattern (sentence beats report, correction beats vanilla) holds on CT despite the shift in modality, encoder backbone, and report style.

\section{Related Work}
\label{sec:related}

\textbf{Medical false-negative correction in VLP.}
We focus on methods that modify the CLIP targets or negatives. MedCLIP~\citep{wang2022medclip} expands paired data combinatorially and replaces InfoNCE with a medical-knowledge-guided semantic matching loss. SAT~\citep{liu2023improving} triages pairs by inter-report similarity. Ko and Park~\citep{ko2025bringing} build dynamic soft labels from text, CheXbert, and RadGraph similarities, combined with negation-based hard negatives. These methods soften image-text pairs through similarity using labels, knowledge, or graphs. CorLIP instead corrects the CLIP targets to assertion-level compatibility: a text is positive when its asserted facts are entailed by the image state, regardless of global report similarity, while the rest of CLIP is left unchanged. Visual-entailment formulations for positive and negative mentions~\citep{wu2025medical} are related; our focus is deriving the contrastive target distribution and using its misspecification to explain both training dynamics.

\textbf{Contrastive-learning theory and supervised contrastive learning.}
General contrastive-learning theory explains when instance discrimination yields semantic representations through latent classes~\citep{saunshi2019theoretical}, alignment and uniformity~\citep{wang2020understanding}, or identifiability of latent generative factors~\citep{zimmermann2021contrastive}. False-negative and soft-target methods relax one-hot contrastive targets by correcting sampled negatives~\citep{chuang2020debiased} or using intra-modal similarity to build softened targets~\citep{gao2024softclip}. These analyses are informative, but radiology lies outside their usual regime: compatible off-diagonal texts are dense rather than rare, reports bundle multiple dependent findings, and many true statements are negative assertions. We therefore view the failure as deterministic target misspecification rather than stochastic class collision. Our mitigation objective is also related to supervised contrastive learning~\citep{khosla2020supervised}, but the relation is not symmetric class equivalence. Radiology compatibility is multi-label, compositional, and directional: an image can entail an assertion without being equivalent to the full report containing it. Unlike a SupCon label set, which defines the entire semantic space, the schema only corrects targets within its coverage; standard CLIP supervision applies elsewhere, preserving open-vocabulary zero-shot inference.

\textbf{Structured reports and noisy labels.}
We use clinical NLP structure only to estimate the contrastive target distribution, drawing on tools for findings, negation, uncertainty, and relations~\citep{peng2018negbio,irvin2019chexpert,smit2020combining,jain2021radgraph}. This differs from prior uses of report structure as classification labels, grounding supervision, or knowledge injection~\citep{huang2021gloria,boecking2022making,liu2024imitate,wu2023medklip}. The rise-then-fall dynamic also resembles noisy-label memorization~\citep{arpit2017closer,zhang2016understanding,xue2022investigating}, but the mechanism is different: the paired label is not randomly corrupted; the one-hot target is structurally incomplete and assigns negative mass to clinically compatible off-diagonal assertions.

\section{Discussion and Limitations}
\label{sec:discussion}

The unit of contrast in radiology should be the assertion, not the report. Reports bundle reusable assertions, so supervision should ask whether an image supports a text assertion, not whether the text is the unique paired report. Sentences are the closest practical approximation to atomic assertions; for bundled reports, even the best soft correction can supply only a single scalar per (image, text) pair, which is lossy when many facts are bundled, leaving the encoder to recover fact-level structure on its own. Sentence-level CorLIP delivers this: across iterations, the same image is queried by different supported assertions, while corrected targets stop repelling compatible off-diagonal texts, encouraging the encoder to encode the image fact state rather than paired-report identity.

Several limits of medical CLIP turn out to be target-specific rather than fundamental. On CXR, large batches once considered harmful now become stable through $B{=}2048$ and deliver useful negatives up to $B{=}512$. Scaling data alone delivers only marginal gains while scaling with corrected targets delivers substantial improvements. The same correction transfers from CXR to 3D CT, indicating the bias is target-level rather than modality-level. Together, these results point to the contrastive target as a key bottleneck of medical CLIP: many practical compromises (small batches, limited data efficiency, modest OOD transfer) stem largely from paired-identity misspecification rather than fundamental constraints. In radiology, defining which pairs count as positive is itself a central design decision, not a fixed convention inherited from general-domain CLIP.

As a design decision, the definition of positives carries a trade-off. CorLIP reads compatibility off a finite schema, so the schema determines which clinical distinctions the target preserves and which it treats as interchangeable. A coarse schema can over-correct: merging left- and right-sided findings, for example, assigns positive target mass to clinically distinct cases, potentially weakening representations of laterality. Removing the incentive to distinguish such cases does not force identical embeddings, but may reduce information useful for finer-grained tasks. Conversely, an increasingly fine flat schema can under-correct: compatible sets become sparse, while related findings receive no shared supervision through their higher-level concepts; in the extreme, compatibility approaches paired identity. Texts with no schema-covered fact retain only their paired positive, so content outside the schema keeps pair-specific supervision. Schema design therefore trades off preserving clinically meaningful distinctions against correcting recurring false-negative conflicts. Our experiments establish benefits for the two evaluated schemas, which differ in both granularity and extractor, rather than identifying an optimal schema.

Our correction is intentionally minimal, and richer compatibility kernels are a natural way to ease this trade-off. Fine-grained assertions over finding, anatomy, laterality, severity, uncertainty, and temporal change, combined through clinical ontologies into hierarchical compatibility, could preserve detailed distinctions while sharing supervision across related findings. Evaluating the corrected representations beyond the image--text similarity probes used here, on report generation, segmentation, and supervised fine-tuning, is an important next step. CorLIP is also orthogonal to other axes of medical CLIP improvement: stronger radiology foundation models and backbones for CXR and CT~\citep{chen2024vision,blankemeier2026merlin,hamamci2026generalist,zhao2025towards}, region-level or global-local objectives~\citep{huang2021gloria,boecking2022making}, specialized clinical knowledge and language models~\citep{wu2023medklip,sellergren2025medgemma}, and larger curated biomedical or CXR corpora~\citep{zhang2023biomedclip,wang2026data}; combining target correction with these advances is a promising direction.

\begin{ack}
This work was supported in part by the Medical Imaging and Data Resource Center (MIDRC), which is funded by the National Institute of Biomedical Imaging and Bioengineering (NIBIB) under contract 75N92020C00021 and through the Advanced Research Projects Agency for Health (ARPA-H). The authors declare no competing interests.
\end{ack}

\bibliographystyle{unsrtnat} 
\bibliography{refs} 

\appendix

\section{Additional Evidence for Target Misspecification and Correction}
\label{app:additional_exp}

\subsection{When Paired-Identity Target Creates False Negatives}
\label{app:problem_visualization}

\begin{figure}[ht]
\centering
\includegraphics[width=0.95\linewidth]{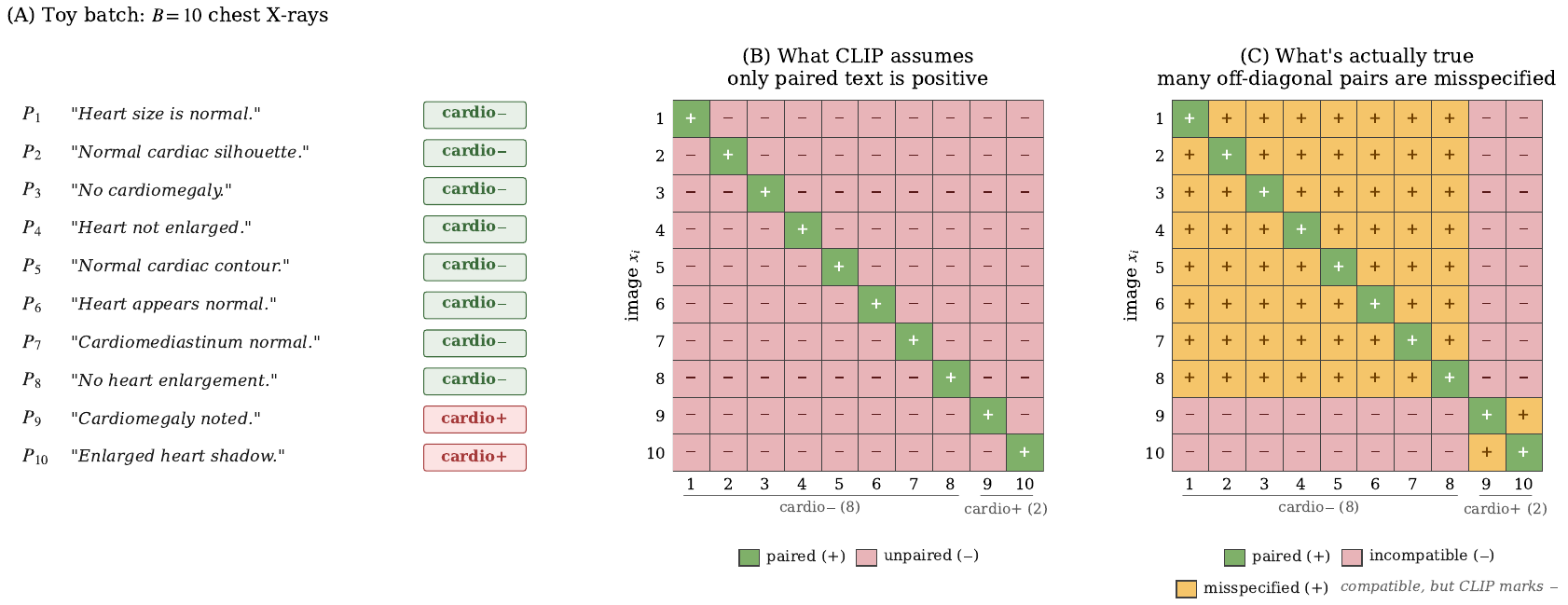}
\caption{\textbf{Target misspecification in medical CLIP.}
(A) A \(B{=}10\) toy batch of CXR reports with \(20\%\) cardiomegaly prevalence; eight reports describe the same normal finding using different paraphrases while two reports describe cardiomegaly with different phrases.
(B) CLIP's paired-identity target marks all 90 off-diagonal image--text pairs as negative.
(C) A compatibility-aware target reveals that 58/90 off-diagonal cells (\(64.4\%\), yellow) are semantically compatible, while the remaining cells (pink) are genuinely incompatible.}
\label{fig:app_problem_visualization}
\end{figure}

Figure~\ref{fig:app_problem_visualization} gives a schematic view of the target mismatch studied in the paper. Vanilla CLIP uses paired identity: in each image-to-text row, only the paired text receives positive mass, and every off-diagonal text is treated as a negative. For radiology image--report data, this row target is too fine-grained. A short assertion can be entailed by many different images, so exact-compatible off-diagonal sentences should not be pure negatives. A longer report may disagree with the anchor image on one finding while still sharing many other asserted facts, so whole-report negatives are often partial positives at the fact level.

The corrected target changes only this supervision matrix. For short assertions, hard correction assigns positive mass to texts exactly entailed by the image. For bundled reports, soft correction assigns mass according to partial fact agreement. This visualization is not an additional method; it is a compact guide to the notation used in Sections~\ref{sec:theory} and~\ref{sec:method}: paired identity \(e_i\), exact compatibility \(K_{ij}\), and partial agreement \(R_{ij}\).

\textbf{False negatives in real MIMIC-CXR batches.} To quantify this on real data, we sample 1{,}000 training batches at $B{=}128$ and $B{=}1024$ and count, for each sentence anchor, the off-diagonal images compatible with it under CheXbert-13d. False negatives arise for both negative and positive assertions. ``No lung opacity'' has $84.9\pm1.2$ false-negative images per batch at $B{=}128$ and $686.3\pm4.6$ at $B{=}1024$ (66.8\% and 67.1\% of off-diagonal entries). ``Lung opacity'' has $13.7\pm0.1$ and $111.2\pm0.4$ (10.8\% and 10.9\%). Even the rare finding ``fracture'' averages $2.1\pm0.1$ and $17.2\pm0.1$ (1.6\% and 1.7\%), and because rare findings already receive weak learning signal, these false negatives carry disproportionate weight.

\subsection{How Compatibility Targets Are Constructed}
\label{app:method_visualization}

\begin{figure}[ht]
\centering
\includegraphics[width=0.95\linewidth]{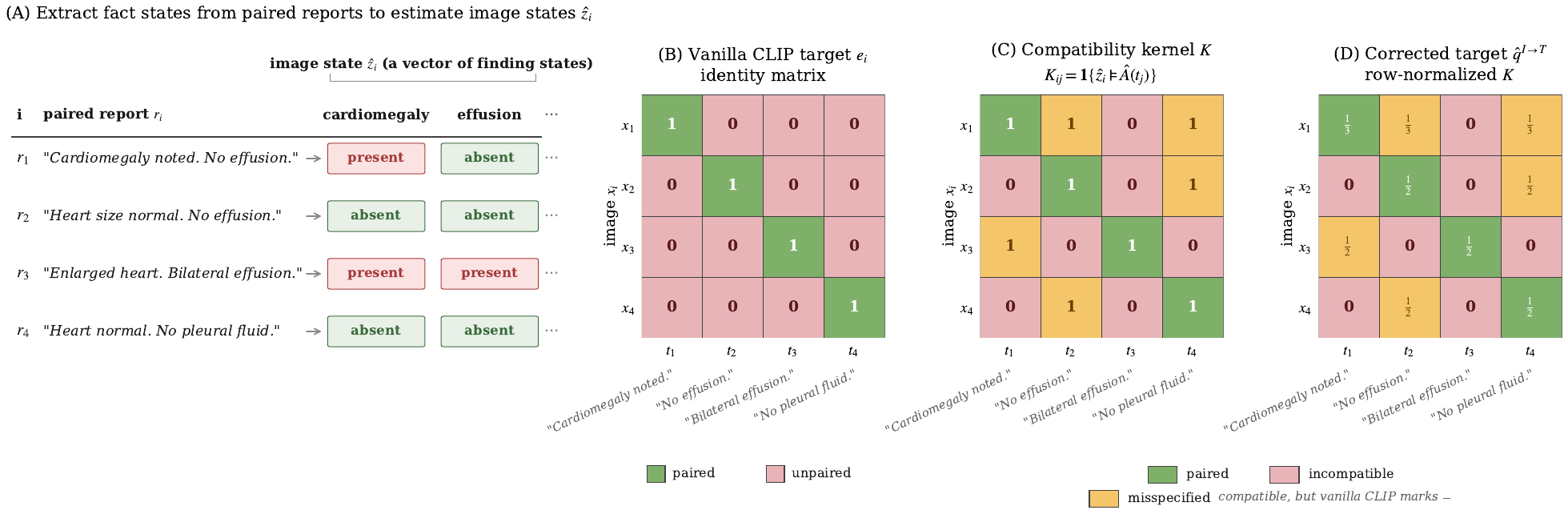}
\caption{\textbf{Constructing corrected targets.}
(A) An offline extractor maps each paired report to an estimated image-level fact state \(\hat z_i\).
(B) Vanilla CLIP uses the identity matrix as its target.
(C) For sampled sentences, the hard compatibility kernel \(K\) marks whether each sentence assertion is entailed by each estimated image state; yellow cells are compatible pairs that vanilla CLIP would mark as negatives.
(D) Row-normalizing \(K\) gives the image-to-text target \(\hat q^{I\to T}\), distributing mass across compatible sentences. The text-to-image target \(\hat q^{T\to I}\) is obtained by column-normalizing \(K\).}
\label{fig:app_method_visualization}
\end{figure}

Figure~\ref{fig:app_method_visualization} illustrates how CorLIP constructs the corrected target in the sentence-level hard-correction case. Since the true image fact state is unobserved, we approximate it by extracting a structured fact vector from the paired full report. The sampled sentence supplies the text-side assertion. The compatibility kernel then records whether each image-side fact state entails each sentence assertion.

Row-normalizing the compatibility kernel gives the image-to-text corrected target. Texts that express the same fact in different phrasings receive the same target support for any image that satisfies that fact, so the correction acts as implicit text augmentation inside the contrastive target. For report and impression correction, the same construction is used with a soft compatibility score: the binary entailment entry is replaced by the fraction of asserted facts satisfied by the image-side state estimate.

\subsection{Target Correction Matters More Than Scaling Data}
\label{app:data_scaling}

Table~\ref{tab:app_data_scaling_mimic} separates two possibilities: vanilla CLIP may fail because the target is biased, or because \(60\mathrm{K}\) training studies are not enough. Doubling the training set to \(120\mathrm{K}\) helps vanilla CLIP only modestly. The best vanilla checkpoint rises from \(68.2\) to \(70.6\) mean AUC, and all three text granularities still degrade after their peak at \(120\mathrm{K}\). Thus the rise-then-fall dynamic is not explained by insufficient data. More data moves the biased trajectory upward, but it does not remove the bias that makes the final checkpoint worse than the peak.

The contrast with target correction is much larger. This diagnostic uses the CheXbert correction. CorLIP-CB trained on only \(60\mathrm{K}\) sentence pairs reaches \(81.6\) peak AUC and \(81.4\) final AUC, already more than 10 points above vanilla CLIP trained on \(120\mathrm{K}\), even when vanilla is evaluated at its own best checkpoint. Target correction also makes additional data more useful: with \(120\mathrm{K}\), corrected sentence-level training reaches \(86.5\) peak and \(86.0\) final AUC, while its peak-to-final drop remains below one point. The lesson is that scale and target quality are not interchangeable. Adding more data to a misspecified target gives limited gains; correcting the target changes the training trajectory and then allows data scaling to pay off.

\begin{table}[ht]
\centering
\caption{\textbf{Data scaling on MIMIC-CXR at \(B{=}128\).} Peak (pk), final (fn), and \(\Delta=\mathrm{fn}-\mathrm{pk}\) MIMIC zero-shot mean AUC, multiplied by 100.}
\label{tab:app_data_scaling_mimic}
\small
\setlength{\tabcolsep}{5.5pt}
\renewcommand{\arraystretch}{1.12}
\begin{tabular}{@{}llrrr rrr@{}}
\toprule
Method & Text & \multicolumn{3}{c}{\(60\mathrm{K}\)} & \multicolumn{3}{c}{\(120\mathrm{K}\)} \\
\cmidrule(lr){3-5}\cmidrule(l){6-8}
& & pk & fn & \(\Delta\) & pk & fn & \(\Delta\) \\
\midrule
Vanilla CLIP & Rep. & 66.1 & 60.1 & $-$6.0 & 68.9 & 63.6 & $-$5.3 \\
Vanilla CLIP & Imp. & 68.1 & 61.1 & $-$7.0 & 69.0 & 61.5 & $-$7.5 \\
Vanilla CLIP & Sen. & 68.2 & 66.5 & $-$1.7 & 70.6 & 67.4 & $-$3.2 \\
\midrule
CorLIP-CB & Rep. & 77.3 & 75.4 & $-$1.9 & 84.5 & 82.1 & $-$2.4 \\
CorLIP-CB & Imp. & 79.0 & 72.5 & $-$6.5 & 84.4 & 80.7 & $-$3.6 \\
CorLIP-CB & Sen. & 81.6 & 81.4 & $-$0.2 & 86.5 & 86.0 & $-$0.6 \\
\bottomrule
\end{tabular}
\end{table}

\subsection{Corrected Targets on \textit{CLEVR-Radiology}}
\label{app:clevr_corrected}

\begin{figure}[ht]
\centering
\includegraphics[width=0.8\linewidth]{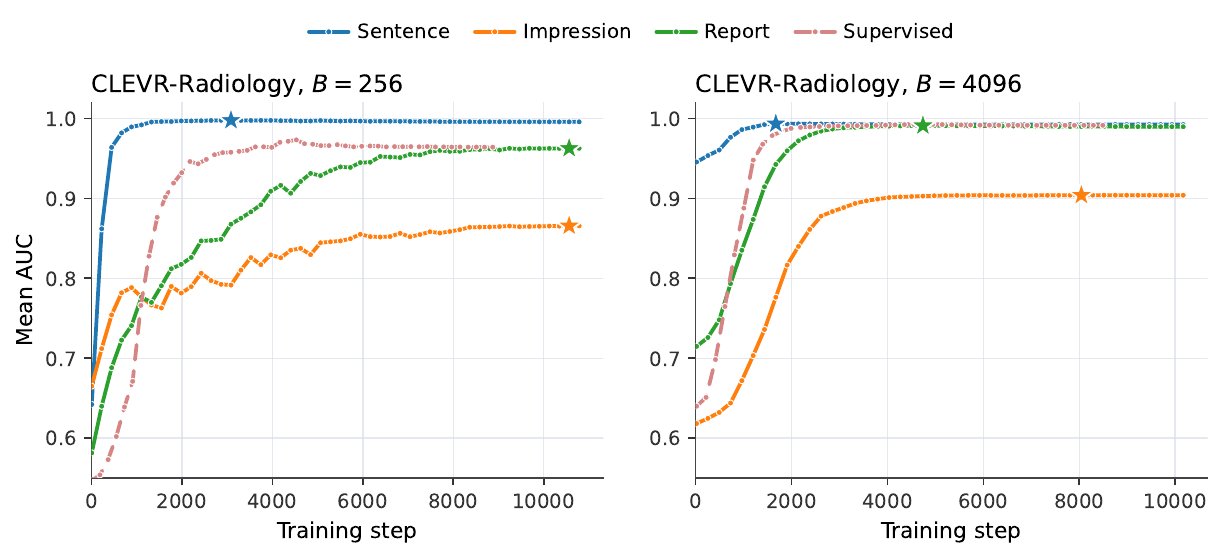}
\caption{\textbf{Corrected targets on \textit{CLEVR-Radiology}.} Evaluation mean AUC at \(B{=}256\) and \(B{=}4096\). Stars mark peak checkpoints; the supervised curve is shown for reference.}
\label{fig:app_clevr_corrected_eval}
\end{figure}

Figure~\ref{fig:app_clevr_corrected_eval} shows the controlled testbed after replacing paired identity with compatibility targets. At both moderate and large batch, corrected training produces stable evaluation trajectories. Sentence-level correction reaches near-perfect AUC fastest and exceeds the supervised classifier at both \(B{=}256\) and \(B{=}4096\); report-level correction also approaches the supervised curve, while impression-level correction remains lower, consistent with impression text being a lossy bundle of multiple facts. This behavior is natural for a corrected contrastive objective: once positives are defined by compatibility, the objective uses many compatible texts per image, much like supervised contrastive learning uses many same-class positives. Prior work shows that supervised contrastive learning can outperform cross-entropy empirically~\citep{khosla2020supervised}, and that supervised contrastive and cross-entropy losses share the same class-collapse simplex geometry at their minima while exhibiting different optimization behavior~\citep{graf2021dissecting}. Our result is the fact-level analogue: corrected contrastive learning does not change the underlying labels in \textit{CLEVR-Radiology}, but it reaches high-quality semantic representations faster by using the compatibility structure during training.

\subsection{Extractor Choice Changes the Tradeoff, Not the Conclusion}
\label{app:extractor_robustness}

Table~\ref{tab:app_extractor_robustness} collects the vanilla and CorLIP rows from Tables~\ref{tab:bs128_indomain} and~\ref{tab:bs128_ood}. The goal is not to rank extractors in isolation, since the two schemas expose different fact sets. Instead, the table checks whether the main conclusion depends on one particular extractor. It does not: both CheXbert and MedGemma correction improve over vanilla CLIP, and both identify sentence-level correction as the most reliable training regime.

The tradeoff follows from the granularity of the extracted fact schema. CheXbert-13d is closely aligned with the CheXpert-style label space used for in-domain MIMIC evaluation, so it gives the strongest MIMIC AUC values. This alignment is also a limitation. The schema is coarse: it covers a small set of findings, does not distinguish laterality, and cannot correct false negatives for findings outside the 13-dimensional label space. For example, left- and right-lung findings can be mapped to the same extracted state, making them compatible under the correction even when localization should distinguish them. MedGemma-30d exposes a broader and finer fact schema, including more findings and laterality. It is therefore less tightly matched to the in-domain MIMIC labels, but transfers better to OOD and dense evaluations, especially VinDr's 28-label classification and MS-CXR grounding. Thus the gains are not explained only by reusing the CheXbert label space: changing the extractor changes which evaluations benefit most, but compatibility correction remains beneficial.

\begin{table}[ht]
\centering
\caption{\textbf{Extractor comparison at \(B{=}128\).} MIMIC reports peak zero-shot mean AUC; CheXpert, VinDr, and MS-CXR report best-checkpoint values. All values are multiplied by 100.}
\label{tab:app_extractor_robustness}
\small
\setlength{\tabcolsep}{5.0pt}
\renewcommand{\arraystretch}{1.12}
\begin{tabular}{@{}lllcccc@{}}
\toprule
Text & Method & Extractor & MIMIC AUC & CheXpert AUC & VinDr AUC & MS-CXR PG \\
\midrule
Rep. & Vanilla & -- & 66.1 & 79.8 & 83.0 & 22.4 \\
Rep. & CorLIP & CheXbert-13d & 77.3 & 84.5 & 83.7 & 34.5 \\
Rep. & CorLIP & MedGemma-30d & 74.1 & 86.2 & \textbf{87.3} & 36.9 \\
\midrule
Imp. & Vanilla & -- & 68.1 & 78.7 & 79.6 & 30.8 \\
Imp. & CorLIP & CheXbert-13d & 79.0 & 81.7 & 82.7 & 33.1 \\
Imp. & CorLIP & MedGemma-30d & 70.8 & 85.9 & 85.2 & 34.1 \\
\midrule
Sen. & Vanilla & -- & 68.2 & 82.1 & 83.1 & 30.7 \\
Sen. & CorLIP & CheXbert-13d & \textbf{81.6} & 85.6 & 83.7 & 37.7 \\
Sen. & CorLIP & MedGemma-30d & 75.8 & \textbf{86.6} & \textbf{87.3} & \textbf{41.6} \\
\bottomrule
\end{tabular}
\end{table}

\subsection{SigLIP Dynamics Under Vanilla and Corrected Targets}
\label{app:siglip}

\begin{figure}[ht]
\centering
\includegraphics[width=0.9\linewidth]{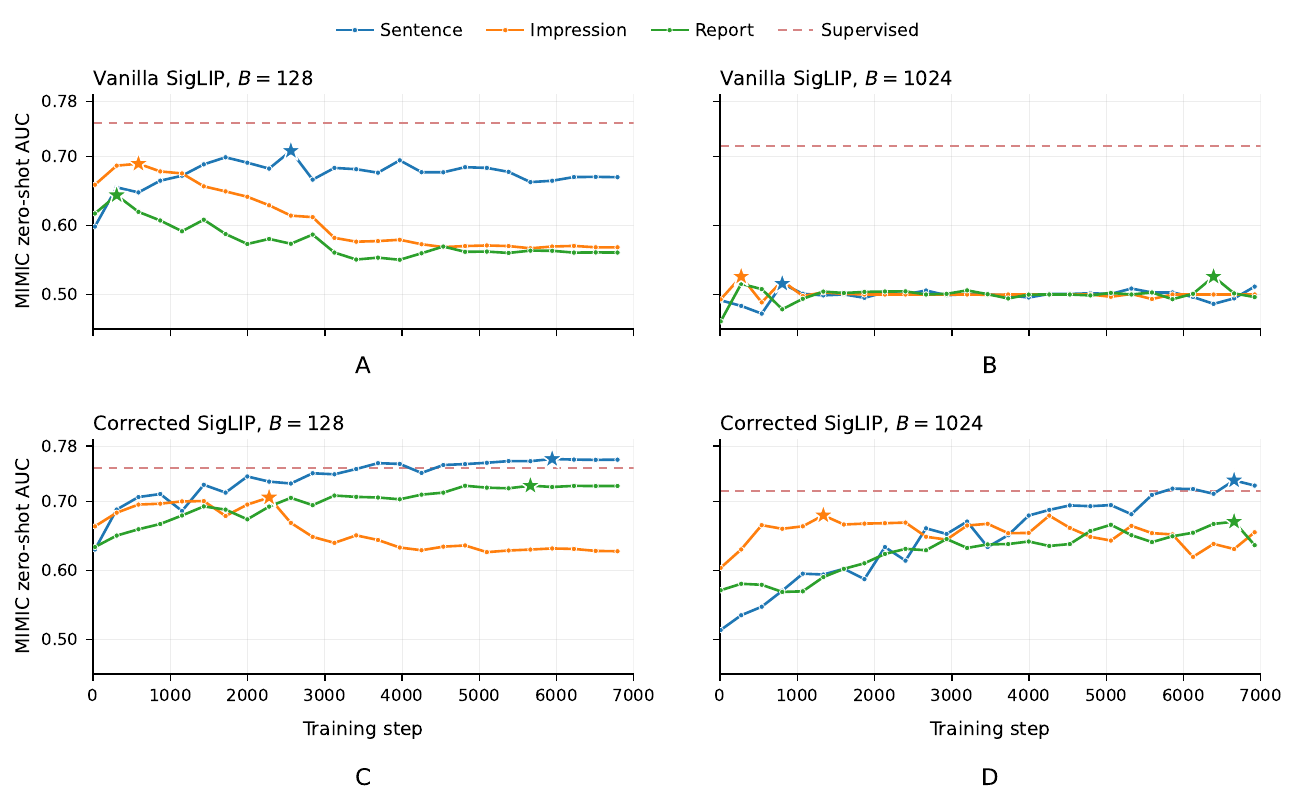}
\caption{\textbf{SigLIP dynamics on MIMIC-CXR.} (A, B) Vanilla SigLIP at \(B{=}128\) and \(B{=}1024\). (C, D) Corrected SigLIP using MedGemma-30d compatibility targets at \(B{=}128\) and \(B{=}1024\). Stars mark peak checkpoints; dashed lines are supervised baselines.}
\label{fig:app_siglip_corrected}
\end{figure}

Figure~\ref{fig:app_siglip_corrected} repeats the MIMIC-CXR dynamics with a SigLIP-style sigmoid contrastive loss~\citep{zhai2023sigmoid}. SigLIP uses the same MIMIC-CXR data, encoders, training schedule, and evaluation protocol as the CLIP experiments. This is a robustness check rather than a full parallel benchmark, but the pattern is informative. Vanilla SigLIP exhibits the same two dynamics as vanilla CLIP: at \(B{=}128\), evaluation rises early and then falls for all three text granularities, even though training continues; at \(B{=}1024\), it collapses to near-random AUC, matching the large-batch failure in Figure~\ref{fig:dynamics}. After replacing paired identity with MedGemma-30d compatibility targets, sentence-level correction becomes stable at both batch sizes and exceeds the supervised baseline in both settings. Thus the correction is not tied to the softmax form of CLIP; it addresses the target structure shared by CLIP and SigLIP-style objectives.

\section{Complete Proofs}
\label{app:proofs}

\subsection{Fact-State Notation and Compatibility}
\label{app:fact_notation}

This appendix expands the notation used in Section~\ref{sec:theory}. Let
\[
    z_i = (z_{i,1},\ldots,z_{i,F})
\]
be the latent fact state of image \(x_i\), where \(a\in\{1,\ldots,F\}\) indexes an atomic finding slot, such as \textsc{cardiomegaly} or \textsc{pleural effusion}, and \(z_{i,a}\) is the value of that slot, such as present or absent. For a text \(t\), let \(\mathcal S(t)\) be the set of fact slots asserted by the text. For each asserted slot \(a\in\mathcal S(t)\), let \(v_a(t)\) be the value asserted by the text. The asserted fact set is
\[
    A(t)=\{(a,v_a(t)):a\in\mathcal S(t)\}.
\]
We say that image \(x_i\) entails text \(t_j\), written \(z_i\models A(t_j)\), if every asserted fact in \(t_j\) agrees with the image fact state:
\[
    z_{i,a}=v_a(t_j)
    \quad \text{for all } a\in\mathcal S(t_j).
\]
The exact compatibility kernel is \(K_{ij}=\mathbf 1\{z_i\models A(t_j)\}\). The paired text is compatible with its image by construction, so \(K_{ii}=1\). The row-normalized exact-compatibility reference is \(q_i^K(j)=K_{ij}/\sum_{\ell=1}^B K_{i\ell}\).

\subsection{Proof of Lemma~\ref{lem:gradient_bias}}
\label{app:proof_lemma_gradient}

\begin{proof}
The proof starts from the standard softmax cross-entropy gradient identity. For any target distribution \(q_i\), \(\nabla_{s_i}\ell_i(q_i) = p_i - q_i\). Indeed, \(\partial \log p_i(j) / \partial s_{ik} = \delta_{jk} - p_i(k)\), so
\[
    \frac{\partial \ell_i(q_i)}{\partial s_{ik}}
    = -\sum_j q_i(j)\bigl[\delta_{jk} - p_i(k)\bigr]
    = p_i(k) - q_i(k),
\]
using \(\sum_j q_i(j) = 1\). Subtracting,
\[
    \nabla_{s_i}\ell_i(q_i) - \nabla_{s_i}\ell_i(\tilde q_i)
    = (p_i - q_i) - (p_i - \tilde q_i)
    = \tilde q_i - q_i.
\]
The model distribution \(p_i\) cancels: the gradient gap depends only on the target mismatch, not on the current model state.

For vanilla CLIP, \(q_i=e_i\), so the logit-space gradient-descent direction is
\[
    -\nabla_{s_i}\ell_i(e_i)=e_i-p_i.
\]
Adding and subtracting any compatibility reference \(q_i^\star\) gives
\[
    -\nabla_{s_i}\ell_i(e_i)
    =
    e_i-p_i
    =
    (q_i^\star-p_i)+(e_i-q_i^\star)
    =
    d_i^{\mathrm{sem}}+d_i^{\mathrm{id}}.
\]
This is Lemma~\ref{lem:gradient_bias}. The first term is the descent direction that would move the row toward the compatibility reference. The second term is the fixed identity residual. In the exact-reference case \(q_i^\star=q_i^K\), this residual has value \(1-1/m_i\) on the paired text, \(-1/m_i\) on each of the other \(m_i-1\) compatible texts, and \(0\) on incompatible texts.
\end{proof}

\noindent\textbf{Immediate consequence: approximate correction sufficiency.}
Let \(q_i^{\mathrm{ref}}\) denote the compatibility reference used for a row: \(q_i^K\) for short exact assertions, or \(q_i^R\) for bundled reports. If \(\hat q_i\) is the practical target used for training, the same softmax identity used in Lemma~\ref{lem:gradient_bias} gives
\[
    \nabla_{s_i}\ell_i(\hat q_i)-\nabla_{s_i}\ell_i(q_i^{\mathrm{ref}})
    =
    q_i^{\mathrm{ref}}-\hat q_i.
\]
Equivalently, the difference between the corresponding descent directions is \(\hat q_i-q_i^{\mathrm{ref}}\). Thus, in logit space, approximate correction improves over vanilla one-hot whenever
\begin{equation}
\label{eq:approx_correction}
    \|\hat q_i - q_i^{\mathrm{ref}}\|_1
    <
    \|e_i - q_i^{\mathrm{ref}}\|_1.
\end{equation}
Parameter-gradient deviations are the corresponding score-Jacobian-weighted versions of these logit-gradient deviations. This is the appendix form of the design principle used in Section~\ref{sec:method}: the closer the constructed target is to the compatibility reference, the smaller the structural gradient bias.

\subsection{Proof of Proposition~\ref{prop:semantic_floor}}
\label{app:proof_prop2}

\begin{proof}
\textbf{Part (i): Semantic floor.}

Let \(P_i=\{j:K_{ij}=1\}\) be the compatible set, with \(|P_i|=m_i\). A purely semantic predictor assigns the same logit \(c_i\) to every \(j\in P_i\). Therefore
\begin{equation}
    p_i(i)
    =
    \frac{\exp(c_i)}
    {m_i\exp(c_i)+\sum_{j\notin P_i}\exp(s_{ij})}
    \leq
    \frac{1}{m_i}.
\end{equation}
The vanilla row loss is \(\ell_i(e_i)=-\log p_i(i)\), hence
\begin{equation}
    \ell_i(e_i)\geq \log m_i.
\end{equation}
Because \(K\) is binary, \(q_i^K\) is uniform over \(P_i\), so \(H(q_i^K)=\log m_i\). The lower bound is approached when all incompatible logits go to \(-\infty\). If the observed vanilla loss drops below \(\log m_i\), the model must be distinguishing among compatible texts using non-semantic features.

\textbf{Part (ii): What happens at \(p_i=q_i^K\).}

The proposition states the behavior at the semantic floor. At \(p_i=q_i^K\), the paired text has probability \(p_i(i)=q_i^K(i)=1/m_i\), so
\[
    \ell_i(e_i)=-\log p_i(i)=\log m_i.
\]
The vanilla descent direction is
\[
    -\nabla_{s_i}\ell_i(e_i)
    =
    e_i-p_i
    =
    e_i-q_i^K
    =
    d_i^{\mathrm{id}}.
\]
Thus the semantic component has vanished, and the only remaining vanilla descent direction is the fixed identity residual.

\textbf{Part (iii): Shortcut incentive.}

Assume $\partial s_{ii}/\partial u = 1$ and $\partial s_{ij}/\partial u = 0$ for $j \neq i$. By the chain rule:
\begin{equation}
    \frac{\partial \ell_i(q_i)}{\partial u} = \frac{\partial \ell_i}{\partial s_{ii}} \cdot \frac{\partial s_{ii}}{\partial u} = p_i(i) - q_i(i),
\end{equation}
using $\partial \ell_i / \partial s_{ii} = p_i(i) - q_i(i)$ from the proof of Lemma~\ref{lem:gradient_bias}.

For vanilla ($q_i(i) = 1$): $\partial \ell_i(e_i)/\partial u = p_i(i) - 1$.

For exact-compatibility-aware training (\(q_i(i) = q_i^K(i)\)): \(\partial \ell_i(q_i^K)/\partial u = p_i(i) - q_i^K(i)\).

At the semantic symmetric point with \(p_i(i) = q_i^K(i) = 1/m_i\):
\begin{equation}
    \frac{\partial \ell_i(e_i)}{\partial u} = \frac{1}{m_i} - 1 = -\!\left(1 - \frac{1}{m_i}\right) < 0, \qquad
    \frac{\partial \ell_i(q_i^K)}{\partial u} = 0.
\end{equation}

Since gradient descent updates \(u \leftarrow u - \eta \cdot \partial\ell/\partial u\), a negative derivative means \(u\) increases; the model is \emph{rewarded} for encoding nuisance features under vanilla training.
\end{proof}

\subsection{Proof of Proposition~\ref{prop:batch_pressure}}
\label{app:proof_prop3}

At initialization \(p_i\approx \pi_B\), where \(\pi_B(j)=1/B\). It is convenient to analyze the vanilla descent direction \(e_i-\pi_B\), since gradient descent on logits moves in the negative gradient direction \(e_i-p_i\). The proof decomposes this direction into a semantic part and an identity part, then compares their squared norms.

\noindent\textbf{Orthogonal decomposition.}
The decomposition
\begin{equation}
    e_i - \pi_B
    \;=\;
    \underbrace{(q_i^K - \pi_B)}_{d_i^{\mathrm{sem}}}
    \;+\;
    \underbrace{(e_i - q_i^K)}_{d_i^{\mathrm{id}}}
\end{equation}
satisfies \(\langle d_i^{\mathrm{sem}}, d_i^{\mathrm{id}} \rangle = 0\).

\noindent\textit{Proof.} Recall \(q_i^K(j) = (1/m_i)\,\mathbf 1\{K_{ij}=1\}\) and \(\pi_B(j) = 1/B\). Expand:
\[
    \langle q_i^K - \pi_B,\; e_i - q_i^K \rangle
    \;=\;
    \langle q_i^K, e_i \rangle - \langle q_i^K, q_i^K \rangle - \langle \pi_B, e_i \rangle + \langle \pi_B, q_i^K \rangle.
\]
The four inner products evaluate as
\begin{align*}
    \langle q_i^K, e_i \rangle &= q_i^K(i) = 1/m_i,
    & \langle q_i^K, q_i^K \rangle &= m_i\cdot(1/m_i)^2 = 1/m_i, \\
    \langle \pi_B, e_i \rangle &= \pi_B(i) = 1/B,
    & \langle \pi_B, q_i^K \rangle &= (1/B)\textstyle\sum_j q_i^K(j) = 1/B,
\end{align*}
giving \(\langle d_i^{\mathrm{sem}}, d_i^{\mathrm{id}} \rangle = 1/m_i - 1/m_i - 1/B + 1/B = 0\). \qed

\noindent\textbf{Component magnitudes.}
Assume \(m_i<B\). If \(m_i=B\), every batch text is compatible, \(q_i^K=\pi_B\), and the semantic component is exactly zero, so the identity-pressure ratio below is infinite.
\begin{equation}
    \|d_i^{\mathrm{sem}}\|_2^2
    \;=\;
    \frac{B - m_i}{m_i\,B},
    \qquad
    \|d_i^{\mathrm{id}}\|_2^2
    \;=\;
    1 - \frac{1}{m_i}.
\end{equation}

\noindent\textit{Derivation.} For \(d_i^{\mathrm{sem}}\), partition the \(B\) coordinates into \(m_i\) compatible texts and \(B-m_i\) incompatible texts:
\begin{align*}
    \|q_i^K - \pi_B\|_2^2
    &= m_i\!\left(\frac{1}{m_i}-\frac{1}{B}\right)^{\!2}
       + (B-m_i)\cdot\frac{1}{B^2} \\
    &= \frac{(B-m_i)^2}{m_i B^2} + \frac{B-m_i}{B^2}
    \;=\; \frac{(B-m_i)\bigl[(B-m_i)+m_i\bigr]}{m_i B^2}
    \;=\; \frac{B-m_i}{m_i B}.
\end{align*}
For \(d_i^{\mathrm{id}}\), the difference \(e_i(j)-q_i^K(j)\) is \(1-1/m_i\) at \(j=i\), \(-1/m_i\) at the other \(m_i-1\) compatible \(j\), and \(0\) at incompatibles, so
\[
    \|e_i - q_i^K\|_2^2
    = \left(1-\frac{1}{m_i}\right)^{\!2} + (m_i-1)\cdot\frac{1}{m_i^2}
    = \frac{(m_i-1)^2 + (m_i-1)}{m_i^2}
    = 1 - \frac{1}{m_i}.
\]

Taking the ratio gives the exact expression in Proposition~\ref{prop:batch_pressure}:
\begin{equation}
    \Gamma_i
    \;=\;
    \frac{\|d_i^{\mathrm{id}}\|_2^2}{\|d_i^{\mathrm{sem}}\|_2^2}
    \;=\;
    \frac{(m_i-1)/m_i}{(B-m_i)/(m_i B)}
    \;=\;
    \frac{B(m_i-1)}{B-m_i}.
\end{equation}

\noindent\textbf{Batch scaling.}
Let \(\omega\) be the probability that an independently sampled off-diagonal text is compatible with the anchor image. Then \(K_{ij}\) for \(j\ne i\) has mean \(\omega\), and \(K_{ii}=1\), so
\[
    \mathbb{E}[m_i]=1+(B-1)\omega\sim B\omega.
\]
Substituting the typical value \(m_i\approx B\omega\) into the exact ratio gives
\begin{equation}
    \Gamma_i
    \;\sim\;
    \frac{\omega}{1-\omega}\,B
    \quad\text{for large }B.
\end{equation}

By orthogonality, \(\|d_i^{\mathrm{van}}\|_2^2 = \|d_i^{\mathrm{sem}}\|_2^2 + \|d_i^{\mathrm{id}}\|_2^2\). Therefore the semantic share of the vanilla descent-direction energy is
\begin{equation}
    \frac{\|d_i^{\mathrm{sem}}\|_2^2}{\|d_i^{\mathrm{van}}\|_2^2}
    \;=\;
    \frac{1}{1+\Gamma_i}
    \;\sim\;
    \frac{1-\omega}{\omega}\cdot\frac{1}{B}
    \quad\text{for large }B,
\end{equation}
which decays as \(1/B\). Intuitively, larger batches add more compatible alternatives. The semantic part still asks the model to bring all compatible texts together, but the identity part asks it to choose the paired text among them; the latter increasingly dominates the vanilla direction as \(B\) grows. \qed

\subsection{Partial positives in bundled reports}
\label{app:report_density}

For bundled reports, exact compatibility \(K\) is conservative: a report can fail \(K_{ij}=1\) on a single contradicted finding while still sharing most asserted facts with the anchor image. The partial-agreement fraction is
\begin{equation}
\label{eq:partial_kernel}
    R_{ij}
    \;=\;
    \frac{1}{|A(t_j)|}
    \sum_{(a,v)\in A(t_j)}
    \mathbf 1\{z_{i,a}=v\}
    \;\in\;[0,1].
\end{equation}
For example, if a candidate report asserts five facts and four agree with the anchor image while one disagrees, then exact compatibility gives \(K_{ij}=0\), but partial agreement gives \(R_{ij}=4/5\). Treating this report as a pure negative would therefore suppress mostly shared facts, not mostly distinguishing facts.

\noindent\textbf{Soft compatibility reference.}
Set \(C_{ii}^R=1\), \(C_{ij}^R=R_{ij}\) for \(j\ne i\), and row-normalize:
\begin{equation}
    q_i^R(j)
    \;=\;
    \frac{C_{ij}^R}{\sum_{\ell=1}^B C_{i\ell}^R}.
\end{equation}
Let \(S_i^R=\sum_{j\ne i}R_{ij}\). Then \(q_i^R(i)=1/(1+S_i^R)\) and \(\sum_{j\ne i}q_i^R(j)=S_i^R/(1+S_i^R)\). Unlike \(q_i^K\), \(q_i^R\) is not an exact-entailment target; it is a soft reference measuring how much shared fact content the one-hot objective assigns the negative sign.

\noindent\textbf{Fact-level sign decomposition.}
The soft score \(R_{ij}\) is useful because it separates shared facts from distinguishing facts. To see this with the simplest algebra, suppose a bundled report score decomposes additively over its asserted facts:
\[
    s_{ij}
    =
    \frac{1}{|A(t_j)|}
    \sum_{(a,v)\in A(t_j)}
    h_i(a,v),
\]
where \(h_i(a,v)\) is the model's score that image \(x_i\) supports fact \((a,v)\). For an off-diagonal report \(j\ne i\), vanilla one-hot has target \(e_i(j)=0\), so
\[
    \frac{\partial \ell_i(e_i)}{\partial s_{ij}}=p_i(j).
\]
Therefore every asserted fact in \(t_j\) receives
\[
    \frac{\partial \ell_i(e_i)}{\partial h_i(a,v)}
    =
    \frac{p_i(j)}{|A(t_j)|},
    \qquad (a,v)\in A(t_j).
\]
Gradient descent lowers all these fact scores. This is right-signed for facts contradicted by \(x_i\), but wrong-signed for facts that are true of \(x_i\). The fraction of asserted facts in \(t_j\) that are true for \(x_i\) is exactly \(R_{ij}\), while the distinguishing fraction is \(1-R_{ij}\). Thus the off-diagonal update carries wrong-signed shared-fact mass \(p_i(j)R_{ij}\) and right-signed distinguishing mass \(p_i(j)(1-R_{ij})\). This is why a report can be a poor negative even when it is not exactly compatible.

\noindent\textbf{Partial-report bias is a deterministic batch-composition function.}

\begin{proposition}[Partial-report bias]
\label{prop:partial_report}
For bundled reports, the deviation between paired identity and the soft compatibility reference is
\begin{equation}
\label{eq:partial_l1}
    \|q_i^R - e_i\|_1
    \;=\;
    2\,\frac{S_i^R}{1+S_i^R},
\end{equation}
which is a deterministic function of the partial-agreement fractions \(\{R_{ij}\}_{j\ne i}\) in the batch. Under independently sampled off-diagonal texts at granularity \(g\),
\begin{equation}
\label{eq:omega_partial}
    \mathbb{E}[S_i^R]
    \;=\;
    (B-1)\rho_g,
    \qquad
    \rho_g \;=\; \mathbb{E}[R(Z,T_g^-)].
\end{equation}
The bias is therefore present in every batch, scales with \(B\), and does not average out with more data or longer training.
\end{proposition}

\noindent\textit{Proof.} Since \(e_i(i)=1\) and \(e_i(j)=0\) for \(j\ne i\),
\[
    \|q_i^R-e_i\|_1
    \;=\;
    \left|q_i^R(i)-1\right|
    + \sum_{j\ne i}q_i^R(j)
    \;=\;
    \left(1 - \frac{1}{1+S_i^R}\right)
    + \frac{S_i^R}{1+S_i^R}
    \;=\;
    \frac{2 S_i^R}{1+S_i^R}.
\]
Equation~\eqref{eq:omega_partial} follows from linearity of expectation applied to \(S_i^R = \sum_{j\ne i} R_{ij}\) using \(\mathbb{E}[R_{ij}] = \rho_g\) for off-diagonal \(j\). \qed

\noindent\textbf{Relation to the exact-compatibility proof.}
The decomposition of Section~\ref{sec:theory:o2} still holds algebraically for the soft reference:
\begin{equation}
\label{eq:decomp_soft}
    e_i - \pi_B
    \;=\;
    \underbrace{(q_i^R - \pi_B)}_{d_i^{\mathrm{sem},R}}
    \;+\;
    \underbrace{(e_i - q_i^R)}_{d_i^{\mathrm{id},R}}.
\end{equation}
When \(R\) is binary, \(q_i^R=q_i^K\), the two components are orthogonal, and Proposition~\ref{prop:batch_pressure} is recovered exactly. For general bundled reports, \(R_{ij}\) is fractional and \(q_i^R\) is not uniform over a compatible set, so the squared-norm ratio depends on the full distribution of partial-agreement scores, not only on their sum \(S_i^R\). We therefore use \(S_i^R\) and \(\|q_i^R-e_i\|_1\) as the clean report-level pressure statistics. They are sufficient for the claim made in Section~\ref{sec:theory:o2}: increasing \(B\) adds more partial-positive mass, and by the softmax target-gap identity that mass is exactly the logit-space bias introduced by paired identity.

\section{Extractor Schemas and Compatibility Construction}
\label{app:compat_edges}

This appendix summarizes how compatibility matrices are instantiated in practice for CXR and CT settings. The main design choice is the fact schema: a coarse schema removes more false negatives by merging related descriptions, while a finer schema preserves clinical distinctions such as laterality, organ system, and device or disease subtype. The schemas below are intentionally simple fixed instantiations of this tradeoff.

\subsection{CXR Fact Schemas}
\label{app:cxr_schema}

\begin{table}[ht]
\centering
\caption{\textbf{Category coverage of the two CXR fact schemas.}}
\label{tab:app_schema_comparison}
\small
\setlength{\tabcolsep}{4.5pt}
\renewcommand{\arraystretch}{1.16}
\begin{tabular}{@{}p{0.20\linewidth}p{0.35\linewidth}p{0.37\linewidth}@{}}
\toprule
Clinical group & \textbf{CheXbert-13d} & \textbf{MedGemma-30d} \\
\midrule
Lung parenchyma
& Lung opacity, lung lesion, edema, consolidation, pneumonia, atelectasis, pneumothorax.
& Right/left opacity, consolidation, atelectasis, edema, pneumothorax, mass or nodule; global vascular congestion, hyperinflation, interstitial markings. \\
\addlinespace[2pt]
Pleura
& Pleural effusion, pleural other.
& Right/left effusion, right/left pleural thickening. \\
\addlinespace[2pt]
Cardiac / mediastinal
& Enlarged cardiomediastinum, cardiomegaly.
& Cardiac enlargement, mediastinal widening, hilar enlargement. \\
\addlinespace[2pt]
Devices
& One support-devices slot.
& Airway tube, vascular line, enteric tube, chest tube, cardiac device. \\
\addlinespace[2pt]
Skeletal / other
& Fracture.
& Fracture, degenerative spine change, diaphragmatic abnormality. \\
\bottomrule
\end{tabular}
\end{table}

Table~\ref{tab:app_schema_comparison} lists only the category coverage. CheXbert~\citep{smit2020combining} is a fixed labeler over the CheXpert-style label space; ``No Finding'' is retained only as auxiliary metadata. MedGemma-27B-Text-IT~\citep{sellergren2025medgemma} is a text-only LLM used as a structured report parser: it extracts anatomy, finding, laterality, and status, which are then resolved into the 30-slot schema above. Sentence-level facts are aggregated over the paired full report to approximate the image-side state \(\hat z_i\). The empirical pattern in Appendix~\ref{app:extractor_robustness} matches the intended tradeoff: CheXbert-13d is strongest when the evaluation label space matches the coarse schema, while MedGemma-30d is stronger when transfer or localization requires finer clinical structure.

\noindent\textbf{MedGemma-30d prompt.}
The prompt defines the finding and anatomy vocabularies of the 30-slot schema and asks for a JSON list of facts per sentence. Its rules extract negated findings as absent, expand broad negations (e.g., ``lungs are clear'') only to findings within their anatomical scope, and keep stable or unchanged findings present. The full prompt is released with our code.

\noindent\textbf{Extraction cost.}
Both extractors are run offline before contrastive training. On a single H100 80GB GPU, CheXbert processes the \(120\mathrm{K}\)-report pool in under half an hour. MedGemma-27B-Text-IT is a larger model, but with vLLM inference acceleration~\citep{kwon2023efficient} it parses the same pool in about \(1.5\) hours. This preprocessing adds no cost during contrastive training or test-time evaluation, and is small relative to the cost of pretraining the vision-language model.

\subsection{CT Fact Schemas}
\label{app:ct_schema}

For the 3D CT transfer experiment, we use a CT-specific MedGemma parser rather than the CXR schemas above. CT reports are commonly organized by organ-system sections, so the parser first segments each report into sections, uses section headers as metadata for implicit anatomy, extracts sentence-level triples \((\text{anatomy}, \text{finding}, \text{laterality}, \text{status})\), and resolves them into the Merlin-30d finding schema used for zero-shot evaluation.

\begin{table}[ht]
\centering
\caption{\textbf{Category coverage of the CT Merlin-30d fact schema.}}
\label{tab:app_ct_schema_comparison}
\small
\setlength{\tabcolsep}{4.5pt}
\renewcommand{\arraystretch}{1.16}
\begin{tabular}{@{}p{0.24\linewidth}p{0.68\linewidth}@{}}
\toprule
Clinical group & \textbf{Merlin-30d categories} \\
\midrule
Lower thorax / cardiac
& Atelectasis, pleural effusion, cardiomegaly, coronary calcification, aortic-valve calcification. \\
\addlinespace[2pt]
Hepatobiliary / pancreas / spleen
& Hepatomegaly, hepatic steatosis, biliary ductal dilation, gallstones, surgically absent gallbladder, pancreatic atrophy, splenomegaly. \\
\addlinespace[2pt]
Renal / adrenal / pelvic organs
& Renal hypodensities, renal cyst, hydronephrosis, prostatomegaly. \\
\addlinespace[2pt]
Vascular / lymph / oncology
& Thrombosis, abdominal aortic aneurysm, atherosclerosis, lymphadenopathy, metastatic disease. \\
\addlinespace[2pt]
GI / peritoneum / abdominal wall / MSK
& Submucosal edema, bowel obstruction, appendicitis, hiatal hernia, ascites, anasarca, free air, fracture, osteopenia. \\
\bottomrule
\end{tabular}
\end{table}

The CT extractor uses MedGemma-27B-Text-IT~\citep{sellergren2025medgemma} as a text-only structured parser. Its section-aware design is important for CT: a sentence such as ``normal size and appearance'' is interpreted differently under liver, kidneys, adrenal glands, or lymph nodes. We therefore use section headers to supply default anatomy for implicit statements and to expand organ-level normal statements into absent findings for the relevant Merlin-30d categories. Present findings override absent findings when sentence-level configs are aggregated into a report-level state. Uncertain triples are not treated as positive assertions in the CT parser. As in CXR, parsing is an offline preprocessing step and MedGemma is not used during contrastive training or evaluation.

\noindent\textbf{Compatibility scoring.}
Across CXR and CT, compatibility is evaluated only on facts asserted by the candidate text \(t_j\). Unmentioned findings in \(t_j\) contribute neither agreement nor disagreement. Sentence-level correction uses hard entailment, \(\hat C_{ij}=1\) if all asserted facts in \(t_j\) are satisfied by \(\hat z_i\), and \(0\) otherwise. Impression- and report-level correction uses soft agreement, the fraction of asserted facts in \(t_j\) satisfied by \(\hat z_i\). In CXR experiments, uncertain findings are mapped to absent/negative; in the CT parser, uncertain triples are ignored. If no clinical fact is extracted from a text unit, we keep only its paired entry positive rather than treating the empty assertion set as compatible with every image. We also force \(\hat C_{ii}=1\) before normalization to guard against extractor inconsistency.

\noindent\textbf{Symmetric targets.}
The same nonnegative compatibility matrix \(\hat C\) is used in both directions. Image-to-text normalizes each row,
\[
    \hat q^{I\to T}_{ij}
    =
    \frac{\hat C_{ij}}{\sum_{\ell=1}^B \hat C_{i\ell}},
\]
and the text-to-image target normalizes each column,
\[
    \hat q^{T\to I}_{ji}
    =
    \frac{\hat C_{ij}}{\sum_{\ell=1}^B \hat C_{\ell j}}.
\]
This is the practical version of the row-wise references \(q_i^K\) and \(q_i^R\) used in the theory; the text-to-image side is the symmetric counterpart.

\noindent\textbf{Scope.}
These schemas are intentionally simple instantiations of the compatibility idea. They use binary or fractional agreement over a fixed set of extracted facts, rather than a learned clinical ontology, severity-aware entailment, uncertainty-aware matching, or a kernel that weights facts by hierarchy and diagnostic importance. The CXR and CT schemas are modality-specific, but the construction is the same: parse reports into asserted facts, estimate the image-side fact state from the paired report, and replace paired identity with compatibility over asserted facts. Designing richer schemas and compatibility kernels is a natural direction for future work, as discussed in Section~\ref{sec:discussion}.

\section{\textit{CLEVR-Radiology}: Construction and Statistics}
\label{app:clevr_data}

This appendix documents the synthetic testbed introduced in Section~\ref{sec:obs}: the generation pipeline, the slot and template inventories used by the semantic-floor computation in Section~\ref{sec:theory:o1}, qualitative samples, dataset statistics, and the floor derivation. The released dataset contains $60{,}000$ image-report pairs, split into $40{,}000/10{,}000/10{,}000$ train/validation/test pairs. Images are rendered at $224\!\times\!224$ and generated by a deterministic, seedable pipeline: scene-config sampling, template-based report rendering, Blender image rendering, and image-noise injection.

\subsection{Generation pipeline}
\label{app:clevr_pipeline}

The pipeline runs in four stages.

\textbf{Stage 1 (scene config).} A scene contains $4$ objects with fixed identities at fixed ground-plane positions: \texttt{left\_cube}, \texttt{right\_cube}, \texttt{sphere}, \texttt{cylinder}. Each object carries $3$ attribute slots (\texttt{color}, \texttt{size}, \texttt{material}), giving $12$ independent slots per scene. Each slot is sampled from an independent Bernoulli: with probability $1{-}p_{\mathrm{abn}}$ the slot takes its normal value (\texttt{gray}, \texttt{medium}, \texttt{rubber} for color, size, material respectively); otherwise an abnormal value is drawn uniformly from that attribute's abnormal vocabulary. Slots are split into $6$ \emph{common} ($p_{\mathrm{abn}}{=}0.20$) and $6$ \emph{rare} ($p_{\mathrm{abn}}{=}0.03$); the assignment is in Table~\ref{tab:clevr_slots}. There are no co-occurrence constraints, so the joint config distribution factorizes across slots.

\textbf{Stage 2 (report).} A template grammar emits a FINDINGS section followed by an IMPRESSION section. FINDINGS iterates over objects and slots: every abnormal slot is mentioned via one of $6$ paraphrase templates; normal slots are mentioned with probability $0.2$ via either an explicit-normal template ($3$ concrete, e.g., ``The sphere is medium in size.''), a positive-abstract template ($3$, e.g., ``The sphere is normal in size.''), or a negative-abstract template ($3$, e.g., ``No abnormal size of the sphere.''). With probability $0.5$ all slots of a fully-normal object are collapsed into a single object-level all-clear sentence ($5$ templates, e.g., ``The sphere is unremarkable.''). The two cubes are merged bilaterally with probability $0.5$ when they share an abnormal value ($4$ bilateral-abnormal templates) or are jointly normal ($5$ bilateral-all-clear templates). IMPRESSION lists abnormal slots joined with commas and ``and'', wrapped in one of $4$ frames (e.g., ``In summary, $\langle$findings$\rangle$.''); fully-normal scenes use one of $6$ all-clear impression templates (``Unremarkable examination.'', ``Normal study.'', etc.). Every emitted sentence keeps its slot tuple metadata $\{(\text{obj}, \text{attr}, \text{value}, \text{is\_abnormal})\}$, so for any (image, sentence) pair the entailment $z_i \models A(t_j)$ is known exactly.

\textbf{Stage 3 (image).} Blender (Cycles, $128$ samples) renders each scene at $224{\times}224$. Object meshes are the standard CLEVR primitives~\citep{johnson2017clevr} (smoothed cube, sphere, smoothed cylinder); colors come from the standard CLEVR palette; size is applied as a scale factor; material is either rubber (default Lambertian) or metal (specular shader). Camera position, three-light rig, and per-object position receive small Gaussian jitter (std $0.15$--$1.0$) seeded by image index for reproducibility. Per-image metadata includes 2D bounding boxes, 3D world coordinates, and pixel-projected centers for each object, supporting downstream grounding evaluation.

\textbf{Stage 4 (image noise).} The rendered CLEVR scenes are visually simple and otherwise contain few image artifacts that could support shortcut learning. To make the visual side less pristine, we add random colored patches after rendering. The number, size, position, and color of these patches are sampled independently for each image and are not tied to the scene config or report text. This perturbation introduces nuisance visual variation while preserving the ground-truth object attributes and fact metadata.

The design mirrors the data-generation structure of radiology without preserving radiology-specific artifacts. The object layout is largely fixed, with small geometric variation, playing the role of shared anatomy. Each object has multiple attribute slots, analogous to finding states over an anatomical region; each slot has only a small value vocabulary, so the semantic space is compositional but not open-ended. The report generator follows radiology-style practice by bundling assertions, always mentioning abnormal states and selectively mentioning normal states. Finally, the random patches introduce image-specific non-semantic variation. Thus \textit{CLEVR-Radiology} retains the shared-schema fact-bundle structure while removing clinical appearance, acquisition artifacts, and radiologist-specific language.

\subsection{Slot table}
\label{app:clevr_slots}

Table~\ref{tab:clevr_slots} lists all $12$ slots with their prior, abnormal vocabulary, and CXR analog. The CXR analog is purely a label for clinical readability; it does not enter any code path or training procedure.

\begin{table}[ht]
\centering
\small
\caption{The $12$ attribute slots of \textit{CLEVR-Radiology}, with two-tier prevalence and CXR analogs.}
\label{tab:clevr_slots}
\begin{tabular}{lllclc}
\toprule
Object & Attribute & Tier & Prior $p_{\mathrm{abn}}$ & Abnormal values & CXR analog \\
\midrule
\texttt{left\_cube}  & \texttt{color}    & common & $0.20$ & red, blue, green & left lung opacity \\
\texttt{left\_cube}  & \texttt{size}     & common & $0.20$ & small, large     & left atelectasis  \\
\texttt{right\_cube} & \texttt{color}    & common & $0.20$ & red, blue, green & right lung opacity \\
\texttt{right\_cube} & \texttt{size}     & common & $0.20$ & small, large     & right atelectasis \\
\texttt{sphere}      & \texttt{size}     & common & $0.20$ & small, large     & cardiomegaly \\
\texttt{cylinder}    & \texttt{color}    & common & $0.20$ & red, blue, green & support devices \\
\midrule
\texttt{left\_cube}  & \texttt{material} & rare   & $0.03$ & metal            & left pleural effusion \\
\texttt{right\_cube} & \texttt{material} & rare   & $0.03$ & metal            & right pleural effusion \\
\texttt{sphere}      & \texttt{color}    & rare   & $0.03$ & red, blue, green & pulmonary edema \\
\texttt{sphere}      & \texttt{material} & rare   & $0.03$ & metal            & pleural other \\
\texttt{cylinder}    & \texttt{size}     & rare   & $0.03$ & small, large     & fracture \\
\texttt{cylinder}    & \texttt{material} & rare   & $0.03$ & metal            & rare finding \\
\bottomrule
\end{tabular}
\end{table}

\subsection{Template inventory}
\label{app:clevr_templates}

Table~\ref{tab:clevr_templates} enumerates the template families and their cardinalities. ``Mention probabilities'' below refers to the per-slot Bernoulli that decides whether a normal slot is mentioned at all, and to the per-object Bernoulli that decides whether all-clear collapsing is used.

\begin{table}[ht]
\centering
\small
\caption{Template families used by the \textit{CLEVR-Radiology} report generator.}
\label{tab:clevr_templates}
\begin{tabular}{@{}llcl@{}}
\toprule
Family & Section & \# templates & Example \\
\midrule
Abnormal mention            & FINDINGS   & $6$ & ``The \{shape\} has \{value\} \{attr\}.'' \\
Normal individual, concrete & FINDINGS   & $3$ & ``The \{shape\} is \{normal\_value\} in \{attr\}.'' \\
Normal individual, positive & FINDINGS   & $3$ & ``The \{shape\} is normal in \{attr\}.'' \\
Normal individual, negative & FINDINGS   & $3$ & ``No abnormal \{attr\} of the \{shape\}.'' \\
Object all-clear            & FINDINGS   & $5$ & ``The \{shape\} is unremarkable.'' \\
Bilateral abnormal (cubes)  & FINDINGS   & $4$ & ``Bilateral \{value\} \{attr\} of the cubes.'' \\
Bilateral all-clear (cubes) & FINDINGS   & $5$ & ``Both cubes are unremarkable.'' \\
Impression abnormal         & IMPRESSION & $4$ & ``In summary, \{findings\}.'' \\
Impression all-normal       & IMPRESSION & $6$ & ``Unremarkable examination.'' \\
\bottomrule
\end{tabular}
\end{table}

Mention probabilities are: normal-mention $0.20$, object-all-clear $0.50$, bilateral-merge $0.50$ (when applicable), bilateral-all-clear $0.40$, concrete-vs-abstract for normals $0.40$, negation-vs-positive for abstract normals $0.30$. These probabilities and template counts are configurable hyperparameters of the report generator, and they are the only inputs (alongside the slot priors) that the floor derivation in Appendix~\ref{app:floor_derivation} consumes.

\subsection{Qualitative samples}
\label{app:clevr_samples}

\begin{figure}[!ht]
\centering
\renewcommand{\arraystretch}{1.6}
\small
\begin{tabular}{@{}m{0.20\linewidth}@{\hspace{1.5em}}m{0.72\linewidth}@{}}
\includegraphics[width=\linewidth]{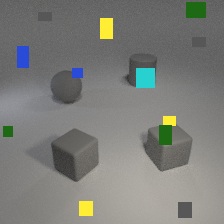} &
(A) \textbf{All-normal.} \texttt{n\_findings=0}. \emph{FINDINGS:} Both cubes are unremarkable. The sphere shows no abnormal color. The cylinder appears normal. \emph{IMPRESSION:} Normal study. \\
\includegraphics[width=\linewidth]{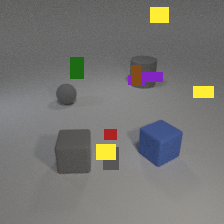} &
(B) \textbf{Two common findings.} \texttt{n\_findings=2}: \texttt{right\_cube.color=blue}, \texttt{sphere.size=small}. \emph{FINDINGS:} The right cube demonstrates blue color. The sphere color is gray. The sphere has small size. The cylinder is normal in color. \emph{IMPRESSION:} blue color of the right cube and small size of the sphere. No other acute abnormality. \\
\includegraphics[width=\linewidth]{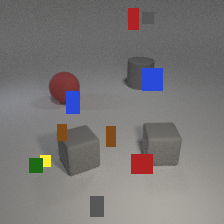} &
(C) \textbf{Rare-only finding.} \texttt{n\_findings=1}: \texttt{sphere.color=red} (a rare slot, $p_{\mathrm{abn}}{=}0.03$). \emph{FINDINGS:} Normal material of the left cube. Normal size of the left cube. No abnormality of the right cube. There is red color involving the sphere. The cylinder appears normal. \emph{IMPRESSION:} red color of the sphere. No other acute abnormality. \\
\includegraphics[width=\linewidth]{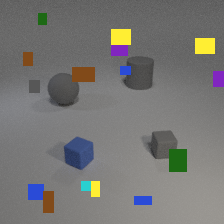} &
(D) \textbf{Bilateral merge.} \texttt{n\_findings=3}: both cubes share \texttt{size=small}, plus \texttt{left\_cube.color=blue}. \emph{FINDINGS:} Both cubes demonstrate small size. The left cube color is blue. No abnormality of the sphere. The cylinder is normal in material. \emph{IMPRESSION:} blue color of the left cube, small size of the left cube and small size of the right cube. \\
\includegraphics[width=\linewidth]{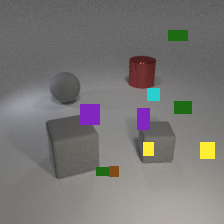} &
(E) \textbf{Multi-finding (common + rare).} \texttt{n\_findings=3}: \texttt{left\_cube.size=large} (common), \texttt{cylinder.color=red} (common), \texttt{cylinder.material=metal} (rare). \emph{FINDINGS:} Large size of the left cube. The left cube material is rubber. The right cube shows no abnormal material. The sphere is medium in size. Metal material of the cylinder. Red color is seen in the cylinder. \emph{IMPRESSION:} large size of the left cube, red color of the cylinder and metal material of the cylinder. No other acute abnormality. \\
\end{tabular}
\caption{Five \textit{CLEVR-Radiology} samples illustrating the dominant report styles. Each row pairs the rendered image with the full FINDINGS and IMPRESSION text.}
\label{fig:clevr_samples}
\end{figure}

\begin{figure}[!ht]
\centering
\renewcommand{\arraystretch}{1.15}
\scriptsize
\begin{tabular}{@{}m{0.13\linewidth}@{\hspace{0.4em}}m{0.32\linewidth}@{\hspace{1.5em}}m{0.13\linewidth}@{\hspace{0.4em}}m{0.32\linewidth}@{}}
\includegraphics[width=\linewidth]{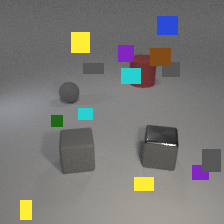} &
\emph{FINDINGS:} The left cube is unremarkable. Metal material of the right cube. Gray color of the right cube. There is small size involving the sphere. The sphere is gray in color. Normal material of the cylinder. The cylinder has red color. \emph{IMPRESSION:} metal material of the right cube, small size of the sphere and red color of the cylinder. No other acute abnormality. &
\includegraphics[width=\linewidth]{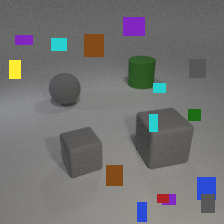} &
\emph{FINDINGS:} The left cube is gray in color. There is large size involving the right cube. The right cube is gray in color. The sphere material is rubber. There is green color involving the cylinder. The cylinder material is unremarkable. \emph{IMPRESSION:} In summary, large size of the right cube and green color of the cylinder. \\[0.4em]

\includegraphics[width=\linewidth]{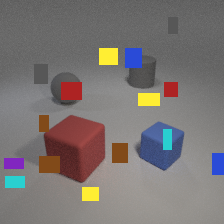} &
\emph{FINDINGS:} Large size of the left cube. The left cube has red color. There is blue color involving the right cube. The sphere is clear. No abnormality of the cylinder. \emph{IMPRESSION:} red color of the left cube, large size of the left cube and blue color of the right cube. &
\includegraphics[width=\linewidth]{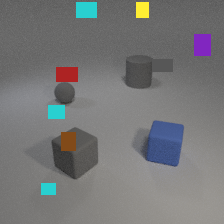} &
\emph{FINDINGS:} The left cube shows no abnormal size. Blue color is seen in the right cube. No abnormal material of the right cube. There is small size involving the sphere. The cylinder appears normal. \emph{IMPRESSION:} Overall, blue color of the right cube and small size of the sphere. \\[0.4em]

\includegraphics[width=\linewidth]{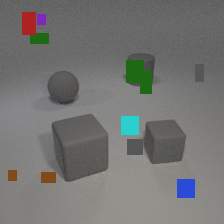} &
\emph{FINDINGS:} The left cube color is unremarkable. The left cube has large size. The right cube appears normal. The sphere material is unremarkable. The cylinder is within normal limits. \emph{IMPRESSION:} In summary, large size of the left cube. &
\includegraphics[width=\linewidth]{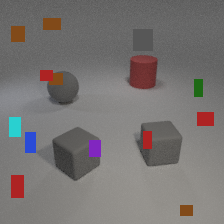} &
\emph{FINDINGS:} The left cube is unremarkable. The sphere shows no abnormal size. No color abnormality of the sphere. Red color is seen in the cylinder. No abnormal material of the cylinder. \emph{IMPRESSION:} Overall, red color of the cylinder. \\[0.4em]

\includegraphics[width=\linewidth]{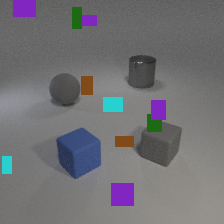} &
\emph{FINDINGS:} Medium size of the left cube. The left cube has blue color. The right cube size is unremarkable. No abnormal size of the sphere. Metal material is seen in the cylinder. \emph{IMPRESSION:} blue color of the left cube and metal material of the cylinder. &
\includegraphics[width=\linewidth]{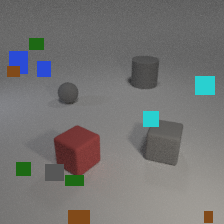} &
\emph{FINDINGS:} Red color is seen in the left cube. Gray color of the right cube. The right cube is normal in size. The sphere has small size. The cylinder is unremarkable. \emph{IMPRESSION:} Overall, red color of the left cube and small size of the sphere. \\[0.4em]

\includegraphics[width=\linewidth]{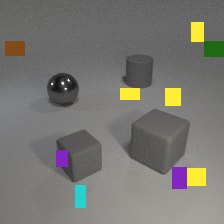} &
\emph{FINDINGS:} Rubber material of the right cube. There is large size involving the right cube. The sphere has metal material. The sphere color is gray. No abnormality of the cylinder. \emph{IMPRESSION:} Overall, large size of the right cube and metal material of the sphere. &
\includegraphics[width=\linewidth]{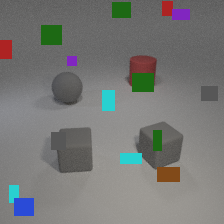} &
\emph{FINDINGS:} The cubes are clear bilaterally. The sphere is clear. The cylinder shows no abnormal size. Red color of the cylinder. \emph{IMPRESSION:} In summary, red color of the cylinder. \\[0.4em]

\includegraphics[width=\linewidth]{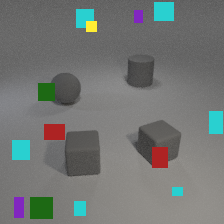} &
\emph{FINDINGS:} The left cube is unremarkable. The right cube is medium in size. The sphere is unremarkable. Rubber material of the cylinder. \emph{IMPRESSION:} Clear. &
\includegraphics[width=\linewidth]{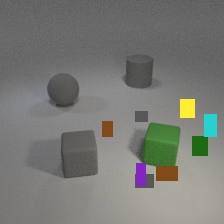} &
\emph{FINDINGS:} The left cube appears normal. There is green color involving the right cube. The sphere is unremarkable. The cylinder is clear. \emph{IMPRESSION:} green color of the right cube. \\[0.4em]

\includegraphics[width=\linewidth]{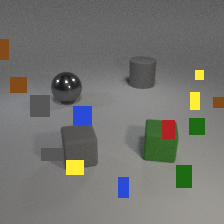} &
\emph{FINDINGS:} The left cube is unremarkable. The right cube demonstrates green color. Metal material is seen in the sphere. The cylinder is gray in color. \emph{IMPRESSION:} Overall, green color of the right cube and metal material of the sphere. &
\includegraphics[width=\linewidth]{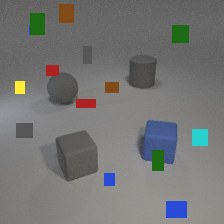} &
\emph{FINDINGS:} The left cube is clear. The right cube demonstrates blue color. No abnormality of the sphere. The cylinder appears normal. \emph{IMPRESSION:} In summary, blue color of the right cube. \\[0.4em]

\includegraphics[width=\linewidth]{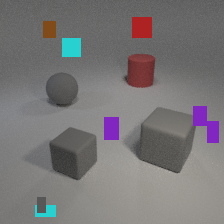} &
\emph{FINDINGS:} Normal color of both cubes. There is large size involving the right cube. The sphere appears normal. The cylinder demonstrates red color. \emph{IMPRESSION:} In summary, large size of the right cube and red color of the cylinder. &
\includegraphics[width=\linewidth]{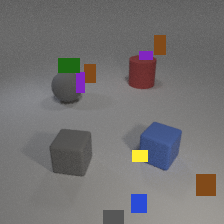} &
\emph{FINDINGS:} There is blue color involving the right cube. The sphere is within normal limits. The cylinder shows no abnormal size. Red color is seen in the cylinder. \emph{IMPRESSION:} blue color of the right cube and red color of the cylinder. \\[0.4em]

\includegraphics[width=\linewidth]{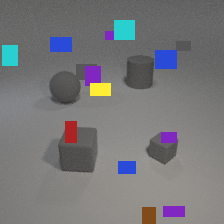} &
\emph{FINDINGS:} Small size is seen in the right cube. The sphere is within normal limits. The cylinder appears normal. \emph{IMPRESSION:} Overall, small size of the right cube. &
\includegraphics[width=\linewidth]{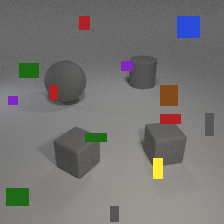} &
\emph{FINDINGS:} The left cube is unremarkable. No abnormality of the right cube. The sphere demonstrates large size. \emph{IMPRESSION:} large size of the sphere. \\[0.4em]

\includegraphics[width=\linewidth]{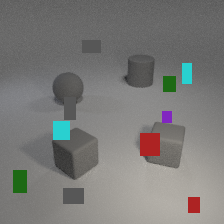} &
\emph{FINDINGS:} The left cube is unremarkable. No abnormal material of the sphere. \emph{IMPRESSION:} No acute findings identified. &
\includegraphics[width=\linewidth]{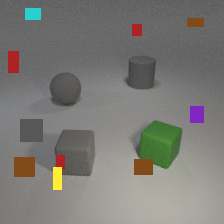} &
\emph{FINDINGS:} The left cube is clear. Green color is seen in the right cube. \emph{IMPRESSION:} Overall, green color of the right cube. \\
\end{tabular}
\caption{Twenty uniformly sampled \textit{CLEVR-Radiology} image-report pairs from the released split (fixed random seed), illustrating the diversity of the generated data.}
\label{fig:clevr_random_samples}
\end{figure}

Figure~\ref{fig:clevr_samples} shows five image-report pairs that span the dominant report styles. The five samples were chosen to cover: an all-normal scene (using object-level and bilateral all-clear templates), a typical $n_{\mathrm{findings}}{=}2$ scene with two common findings, a rare-only finding (a red sphere, where ``red'' is rare for the sphere's color slot), a bilateral-merged cube finding, and a multi-finding scene mixing common and rare slots. Additional dataset samples are in Figure~\ref{fig:clevr_random_samples}.

\subsection{Additional random samples}
\label{app:clevr_random}

\subsection{Dataset statistics}
\label{app:clevr_stats}

Table~\ref{tab:clevr_stats} summarizes the released split. Three numbers are worth highlighting for the theory: (i) report texts are nearly all unique ($94.6\%$), so identity-target training has near-zero exact text duplication, yet (ii) the underlying \emph{config} space collapses to $3{,}027$ distinct configs and the \emph{binary abnormality pattern} space to only $707$ patterns, meaning many ``different'' reports describe the same finding set. This gap between text-level distinctness and fact-level distinctness is exactly the regime where the semantic floor of Section~\ref{sec:theory:o1} becomes load-bearing. (iii) The all-normal rate observed at $21.9\%$ matches the closed-form prediction $(1{-}0.20)^{6}\!\cdot\!(1{-}0.03)^{6}\!\approx\!21.8\%$ to within sampling noise, a sanity check that the generator and the prior tabulation agree.

\begin{table}[ht]
\centering
\small
\caption{\textit{CLEVR-Radiology} dataset statistics over the $60{,}000$ released image-report pairs.}
\label{tab:clevr_stats}
\begin{tabular}{lr}
\toprule
\textbf{Scale} & \\
Image-report pairs                                        & $60{,}000$ \\
Image format / resolution                                 & PNG, $224\!\times\!224$ \\
\midrule
\textbf{Diversity} & \\
Unique report texts                                       & $56{,}751$ \\
Unique configs (exact attribute values)                   & $3{,}027$ \\
Unique binary abnormality patterns                        & $707$ \\
Unique sentence texts                                     & $6{,}202$ \\
\midrule
\textbf{Report length (words)} & \\
Mean / median / range                                     & $35$ / $34$ / $7$--$99$ \\
Sentences per report (mean / median / range)              & $5.0$ / $5$ / $2$--$11$ \\
\midrule
\textbf{Finding count distribution} & \\
$n_{\mathrm{findings}}{=}0$ (all-normal)                  & $21.9\%$ \\
$n_{\mathrm{findings}}{=}1$                               & $37.0\%$ \\
$n_{\mathrm{findings}}{=}2$                               & $26.8\%$ \\
$n_{\mathrm{findings}}{=}3$                               & $11.0\%$ \\
$n_{\mathrm{findings}}{\geq}4$                            & $\phantom{0}3.4\%$ \\
\bottomrule
\end{tabular}
\end{table}

\subsection{Generator-based semantic floor}
\label{app:floor_derivation}

By Proposition~\ref{prop:semantic_floor}, the row-$i$ semantic floor of vanilla CLIP is \(\log m_i=H(q_i^K)\), where \(q_i^K\) is the exact-compatibility reference over the \(B\) batch texts. For \textit{CLEVR-Radiology} the generator records the full image config and the asserted fact set of every emitted text, so the compatibility probability can be computed directly from generator metadata. We summarize the report-level and sentence-level calculations used for the dashed lines in Figure~\ref{fig:theory_evidence}(A, B).

\noindent\textbf{Computation.}
For a fixed anchor, the floor depends only on how many other batch texts are compatible with it. We therefore estimate, under the generator distribution, the number of compatible off-diagonal texts in a \(B\)-sized batch and take the expected \(\log\) positive count after adding the paired text itself. This is the same generator-based calculation used for the dashed lines in Figure~\ref{fig:theory_evidence}(A, B).

\noindent\textbf{Report level.}
For report-level training, compatibility is computed from the complete image config and the asserted fact set of a generated report. The config prior is known exactly from the slot sampling probabilities in Table~\ref{tab:clevr_slots}, so the report floor is computed by enumerating the config prior and evaluating how often independently generated reports are entailed by each config. This gives the report-level dashed floor without using any model outputs.

\noindent\textbf{Sentence level.}
For sentence-level training, the relevant object is a sentence pattern: the set of slot-value facts asserted by one generated sentence. We estimate the distribution of these patterns using the same generator defaults as the released data, including normal-mention probability, object-level all-clear sentences, bilateral merges, and the rule that training samples one sentence from a report. The sentence floor then uses the probability that a random sentence pattern is entailed by a random image config, averaged symmetrically over image-to-text and text-to-image directions.

\noindent\textbf{Use in Figure~\ref{fig:theory_evidence}.}
The dashed floors are precomputed from the slot priors, template probabilities, emitted assertion metadata, and batch size \(B{=}256\). They are not fit to the training loss. The training curves are used only afterward to check whether the observed loss at the eval-peak step lies near the generator-predicted semantic floor.

\section{Experimental Setup Details}
\label{app:exp}

\subsection{CXR Experiments}
\label{app:exp_cxr}

\textbf{Data.}
MIMIC-CXR contains many image views and report formats. To keep the study focused on the contrastive target, we restrict pretraining to frontal images and reports containing both FINDINGS and IMPRESSION sections. This filtering yields a training pool slightly above \(120\mathrm{K}\) image--report pairs. Because MS-CXR is annotated on top of MIMIC-CXR, we remove all MS-CXR images from the pretraining pool to avoid overlap with the grounding evaluation. We use two training scales, \(60\mathrm{K}\) and \(120\mathrm{K}\), and keep the official validation and test splits for validation and in-domain evaluation.

\textbf{Text units.}
Report uses FINDINGS+IMPRESSION, impression uses IMPRESSION alone, and sentence samples one report sentence per training iteration.

\textbf{Evaluation data.}
CheXpert and VinDr are used only for zero-shot out-of-distribution classification, never for training. We evaluate on their official labeled test splits: CheXpert uses the same 14-label disease vocabulary as MIMIC-CXR, while VinDr provides a finer 28-label CXR vocabulary. For dense evaluation, we use MS-CXR phrase-grounding annotations, which pair radiology finding phrases with radiologist-verified bounding boxes. Since MS-CXR is small, we use all annotated examples for testing.

\textbf{Evaluation protocol.}
For zero-shot classification, all methods, including false-negative mitigation baselines, use the same class-name prompts and scoring rule. For phrase grounding, we compute a phrase heatmap by taking the similarity between the text embedding of the queried phrase and the dense image embeddings, then resizing the resulting similarity map to the input image size. We compute pointing-game accuracy on MS-CXR: a prediction is counted as correct if the maximum value of the phrase heatmap falls inside any ground-truth bounding box for that phrase.

For in-domain MIMIC-CXR training-dynamics analyses, ``peak'' denotes the logged checkpoint with the highest test-set metric and ``final'' denotes the last checkpoint. We use this retrospective peak only as a diagnostic, not as a deployable model-selection rule: the goal is to measure how far the final checkpoint falls from the best representation reached along the trajectory, and the full trajectory is shown whenever this comparison is central. For OOD classification and grounding tables, checkpoints are selected by validation performance and the corresponding test-set results are reported.

\textbf{CXR training.}
All CXR methods are trained in the same framework. Vanilla CLIP, mitigation baselines, and CorLIP variants share the same DINOv3-initialized ViT-B/16 image encoder~\citep{simeoni2025dinov3,dosovitskiy2020image}, BioClinicalBERT text encoder~\citep{devlin2019bert,alsentzer2019publicly}, \(224\times224\) image resolution, augmentations, optimizer, schedule, and checkpointing. They differ only in the contrastive target used for each batch. We use AdamW~\citep{loshchilov2017decoupled} with default momentum parameters and weight decay \(10^{-4}\). The base learning rate is \(10^{-4}\) at \(B{=}128\); other batch sizes use batch-size-adapted learning rates capped at \(4\times10^{-4}\). The schedule uses linear warmup over the first \(5\%\) of steps followed by cosine decay. Augmentations include random crop, brightness jitter, and Gaussian noise.

\textbf{Baseline rationale.}
The vanilla CLIP rows at report, impression, and sentence granularity already instantiate the paired-identity objectives underlying several classic CXR VLP systems. Report- or impression-level paired training corresponds to CheXzero-style image--report CLIP~\citep{tiu2022expert}, while sentence-level paired training corresponds to ConVIRT-style sentence sampling~\citep{zhang2022contrastive}. We do not reuse public checkpoints because they differ in data filtering, encoder initialization, optimization, image processing, and training scale; instead, we reinstantiate the paired-identity objective inside our controlled pipeline so that the target is the only intended difference. For false-negative mitigation baselines, we reimplement SAT~\citep{liu2023improving}, MedCLIP~\citep{wang2022medclip}, and Ko\&Park~\citep{ko2025bringing} in the same codebase. Method-specific target construction and hyperparameters follow the original papers, while all shared training choices follow the CXR setup above. These baselines are not incompatible with our framework; they can be viewed as alternative choices of text granularity and target kernel, using report similarity, semantic matching, graph similarity, or dynamic soft labels instead of explicit fact compatibility. CorLIP isolates the compatibility kernel predicted by our theory while keeping the rest of the training pipeline fixed.

We do not compare against every radiology VLP system that improves performance through an orthogonal axis. Methods such as GLoRIA~\citep{huang2021gloria} and BioViL~\citep{boecking2022making} add global-local or dense alignment objectives; MedKLIP~\citep{wu2023medklip} and IMITATE~\citep{liu2024imitate} inject medical knowledge or clinical priors; larger foundation models change data scale, backbones, or language modeling capacity. These choices are valuable, but they test a different axis from the one studied here. CorLIP is meant to be a target-level correction and can in principle be combined with stronger architectures, local objectives, or larger pretraining corpora.

\textbf{Supervised baseline.}
The supervised control uses the same CXR training data, image backbone, augmentations, optimizer, schedule, and checkpointing as the contrastive runs. The only architectural change is replacing the contrastive projection with a multi-label classification head. We train this head with supervised cross-entropy using the CheXpert 14-class labels provided with MIMIC-CXR~\citep{johnson2019mimic,irvin2019chexpert}. This baseline serves only as a non-contrastive control for the training dynamics in Observation~1.

\subsection{\textit{CLEVR-Radiology} Experiments}
\label{app:exp_clevr}

For \textit{CLEVR-Radiology}, we use an OpenCLIP ViT-B/32 image encoder~\citep{cherti2023reproducible} with random initialization, paired with the same BERT-style text encoder family unless otherwise stated. The image resolution, optimizer, schedule, and augmentation recipe follow the CXR setup. Dataset construction details are in Appendix~\ref{app:clevr_data}.

\subsection{3D CT Transfer Experiment}
\label{app:exp_ct}

To test whether target correction transfers beyond 2D CXR, we run a supporting experiment on the Merlin 3D CT image--report pretraining setup~\citep{blankemeier2026merlin}. The training set contains \(25{,}528\) CT scans paired with radiology reports. Following Merlin, we use the FINDINGS section as the text source, since it contains detailed organ-system observations. We evaluate three configurations while keeping image inputs, preprocessing, encoders, optimizer, and schedule fixed: full-report vanilla contrastive training, random sentence-level vanilla contrastive training, and sentence-level CorLIP, for which MedGemma-27B-Text-IT~\citep{sellergren2025medgemma} is used offline as a structured text parser to extract CT finding assertions and construct sentence-level compatibility targets; it is not used during model training or evaluation.

The CT image encoder is an I3D ResNet-152 and the text encoder is a Clinical Longformer, with both encoders producing \(512\)-dimensional embeddings. We train for \(52\) epochs using the same CT image preprocessing pipeline as Merlin. Optimization uses AdamW with \(\beta=(0.9,0.999)\), initial learning rate \(10^{-5}\), and a cosine learning-rate schedule parameterized over \(300\) epochs. Gradient checkpointing is applied to both encoders, and training uses FP16 mixed precision. These settings allow a maximum batch size of \(18\) on a single NVIDIA L40S 48GB GPU; evaluation reports zero-shot classification over \(30\) Merlin findings as mean AUC with \(95\%\) confidence intervals.

\subsection{Compute Resources}
\label{app:exp_compute}

CXR and \textit{CLEVR-Radiology} experiments were run with BF16 precision on NVIDIA H100 80GB GPUs on nodes with approximately 500GB system memory. For report-level CXR training, each GPU supports a batch size of 128. For sentence- and impression-level training, the shorter text units allow a batch size of 256 per GPU, so the number of GPUs used for each global-batch experiment can be inferred from the reported batch size and the corresponding per-GPU capacity. The Merlin CT experiment was run separately with FP16 precision on a single NVIDIA L40S 48GB GPU because of the data storage requirements.

\newpage
\section*{NeurIPS Paper Checklist}

\begin{enumerate}

\item {\bf Claims}
    \item[] Question: Do the main claims made in the abstract and introduction accurately reflect the paper's contributions and scope?
    \item[] Answer: \answerYes{} 
    \item[] Justification: Main claims are stated in the abstract and introduction sections.

\item {\bf Limitations}
    \item[] Question: Does the paper discuss the limitations of the work performed by the authors?
    \item[] Answer: \answerYes{} 
    \item[] Justification: Limitations are discussed in Section~\ref{sec:discussion}

\item {\bf Theory assumptions and proofs}
    \item[] Question: For each theoretical result, does the paper provide the full set of assumptions and a complete (and correct) proof?
    \item[] Answer: \answerYes{} 
    \item[] Justification: Theoretical assumptions are stated in Section~\ref{sec:theory}, with complete proofs and notation in Appendix~\ref{app:proofs}.

    \item {\bf Experimental result reproducibility}
    \item[] Question: Does the paper fully disclose all the information needed to reproduce the main experimental results of the paper to the extent that it affects the main claims and/or conclusions of the paper (regardless of whether the code and data are provided or not)?
    \item[] Answer: \answerYes{} 
    \item[] Justification: Datasets, text granularities, training setup, baselines, evaluation protocols, and hyperparameters are described in Section~\ref{sec:exp} and Appendix~\ref{app:exp}.

\item {\bf Open access to data and code}
    \item[] Question: Does the paper provide open access to the data and code, with sufficient instructions to faithfully reproduce the main experimental results, as described in supplemental material?
    \item[] Answer: \answerYes{} 
    \item[] Justification: The data used in this paper are all publicly available, and the code to reproduce the main experimental results is publicly available. Instructions for reproducing the main experimental results are provided in Appendix~\ref{app:exp}.

\item {\bf Experimental setting/details}
    \item[] Question: Does the paper specify all the training and test details (e.g., data splits, hyperparameters, how they were chosen, type of optimizer) necessary to understand the results?
    \item[] Answer: \answerYes{} 
    \item[] Justification: Training data, preprocessing, model initialization, optimizer, schedule, augmentations, baselines, and evaluation details are provided in Appendix~\ref{app:exp}.

\item {\bf Experiment statistical significance}
    \item[] Question: Does the paper report error bars suitably and correctly defined or other appropriate information about the statistical significance of the experiments?
    \item[] Answer: \answerNo{} 
    \item[] Justification: We do not report multi-seed error bars or formal significance tests. The study reimplements all methods in a shared training framework for controlled comparison, and sweeps over many text granularities, batch sizes, data scales, extractors, and evaluation datasets; repeating the full grid across multiple seeds would exceed our compute budget. To make the conclusions inspectable, we report complete training trajectories, peak/final checkpoints, and consistent trends across granularities, datasets, and batch sizes.

\item {\bf Experiments compute resources}
    \item[] Question: For each experiment, does the paper provide sufficient information on the computer resources (type of compute workers, memory, time of execution) needed to reproduce the experiments?
    \item[] Answer: \answerYes{} 
    \item[] Justification: The experiments compute is described in Appendix~\ref{app:exp}
    
\item {\bf Code of ethics}
    \item[] Question: Does the research conducted in the paper conform, in every respect, with the NeurIPS Code of Ethics \url{https://neurips.cc/public/EthicsGuidelines}?
    \item[] Answer: \answerYes{} 
    \item[] Justification: This work follows the NeurIPS Code of Ethics. We have reviewed the Code of Ethics and confirm that our research conforms to it in every respect.

\item {\bf Broader impacts}
    \item[] Question: Does the paper discuss both potential positive societal impacts and negative societal impacts of the work performed?
    \item[] Answer: \answerYes{} 
    \item[] Justification: Section~\ref{sec:discussion} discusses potential benefits for medical representation learning.
    
\item {\bf Safeguards}
    \item[] Question: Does the paper describe safeguards that have been put in place for responsible release of data or models that have a high risk for misuse (e.g., pre-trained language models, image generators, or scraped datasets)?
    \item[] Answer: \answerNA{} 
    \item[] Justification: The paper does not release a high-risk generative model or scraped dataset.

\item {\bf Licenses for existing assets}
    \item[] Question: Are the creators or original owners of assets (e.g., code, data, models), used in the paper, properly credited and are the license and terms of use explicitly mentioned and properly respected?
    \item[] Answer: \answerYes{} 
    \item[] Justification: Existing datasets, models, and tools are cited.

\item {\bf New assets}
    \item[] Question: Are new assets introduced in the paper well documented and is the documentation provided alongside the assets?
    \item[] Answer: \answerYes{} 
    \item[] Justification: The synthetic dataset construction, statistics, templates, and examples are documented in Appendix~\ref{app:clevr_data}; the dataset is released together with the code.

\item {\bf Crowdsourcing and research with human subjects}
    \item[] Question: For crowdsourcing experiments and research with human subjects, does the paper include the full text of instructions given to participants and screenshots, if applicable, as well as details about compensation (if any)? 
    \item[] Answer: \answerNA{} 
    \item[] Justification: The work does not conduct new crowdsourcing or new human-subject experiments; it uses existing de-identified datasets and annotations.

\item {\bf Institutional review board (IRB) approvals or equivalent for research with human subjects}
    \item[] Question: Does the paper describe potential risks incurred by study participants, whether such risks were disclosed to the subjects, and whether Institutional Review Board (IRB) approvals (or an equivalent approval/review based on the requirements of your country or institution) were obtained?
    \item[] Answer: \answerNA{} 
    \item[] Justification: The study uses existing de-identified public datasets and does not collect new human-subject data.

\item {\bf Declaration of LLM usage}
    \item[] Question: Does the paper describe the usage of LLMs if it is an important, original, or non-standard component of the core methods in this research? Note that if the LLM is used only for writing, editing, or formatting purposes and does \emph{not} impact the core methodology, scientific rigor, or originality of the research, declaration is not required.
    \item[] Answer: \answerYes{} 
    \item[] Justification:  MedGemma-27B-Text-IT is used as a structured report parser for one extractor schema, as described in Appendix~\ref{app:compat_edges}; LLM use for writing/editing is not part of the scientific method.

\end{enumerate}

\end{document}